\documentclass{article}

\usepackage{microtype}
\usepackage{graphicx}
\usepackage{booktabs} 

\usepackage{hyperref}

\usepackage{amsmath}
\usepackage{amssymb}
\usepackage{amsthm}
\usepackage{caption}
\usepackage{subcaption}
\usepackage{graphicx}
\usepackage[dvipsnames]{xcolor}
\usepackage[symbol]{footmisc}

\DeclareFontFamily{OT1}{pzc}{}
\DeclareFontShape{OT1}{pzc}{m}{it}{<-> s * [1.10] pzcmi7t}{}
\DeclareMathAlphabet{\mathpzc}{OT1}{pzc}{m}{it}

\usepackage{tabularx}
\usepackage{booktabs} 

\newtheorem{theorem}{Theorem}[section]

\newtheorem{lemma}[theorem]{Lemma}
\theoremstyle{definition}

\newtheorem{innercustomthm}{Theorem}
\newenvironment{customthm}[1]
  {\renewcommand\theinnercustomthm{#1}\innercustomthm}
  {\endinnercustomthm}

\newtheorem{innercustomlemma}{Lemma}
\newenvironment{customlemma}[1]
  {\renewcommand\theinnercustomlemma{#1}\innercustomlemma}
  {\endinnercustomlemma}

\newcommand{\bE}{\boldsymbol{\mathrm{E}}}

\newcommand{\bx}{\boldsymbol{x}}

\newcommand{\bv}{\boldsymbol{\mathrm{v}}}

\newcommand{\bW}{\boldsymbol{\mathrm{W}}}
\newcommand{\bw}{\boldsymbol{\mathrm{w}}}

\newcommand{\btheta}{\boldsymbol{\theta}}
\newcommand{\bH}{\boldsymbol{\mathrm{H}}}
\newcommand{\bG}{\boldsymbol{\mathrm{G}}}
\newcommand{\bg}{\boldsymbol{\mathrm{g}}}
\newcommand{\bI}{\boldsymbol{\mathrm{I}}}
\newcommand{\bsig}{\boldsymbol{\sigma}}
\newcommand{\bSigma}{\boldsymbol{\Sigma}}
\newcommand{\bet}{\boldsymbol{\eta}}
\newcommand{\ba}{\boldsymbol{\mathrm{a}}}
\newcommand{\bA}{\boldsymbol{\mathrm{A}}}
\newcommand{\bs}{\boldsymbol{s}}
\newcommand{\bS}{\boldsymbol{S}}
\newcommand{\trace}{\mathrm{tr}}

\newcommand{\bmu}{\boldsymbol{\mu}}

\newcommand{\bLambda}{\boldsymbol{\Lambda}}

\DeclareMathOperator*{\argmax}{arg\,max}

\usepackage[accepted]{icml2020}

\icmltitlerunning{Dissecting Non-Vacuous Generalization Bounds}

\begin{document}

\twocolumn[
\icmltitle{Dissecting Non-Vacuous Generalization Bounds\\
           based on the Mean-Field Approximation}



\icmlsetsymbol{equal}{*}

\begin{icmlauthorlist}
\icmlauthor{Konstantinos Pitas}{to}
\end{icmlauthorlist}

\icmlaffiliation{to}{Institute of Electrical Engineering, EPFL, Lausanne, Switzerland}

\icmlcorrespondingauthor{Konstantinos Pitas}{konstantinos.pitas@epfl.ch}

\icmlkeywords{Machine Learning, ICML}

\vskip 0.3in
]



\printAffiliationsAndNotice{}  

\begin{abstract}
Explaining how overparametrized neural networks simultaneously achieve low risk and zero empirical risk on benchmark datasets is an open problem. PAC-Bayes bounds optimized using variational inference (VI) have been recently proposed as a promising direction in obtaining non-vacuous bounds. We show empirically that this approach gives negligible gains when modeling the posterior as a Gaussian with diagonal covariance---known as the mean-field approximation. We investigate common explanations, such as the failure of VI due to problems in optimization or choosing a suboptimal prior. Our results suggest that investigating richer posteriors is the most promising direction forward.
\end{abstract}

\section{Introduction}
Two recent works \citet{dziugaite2017computing,zhou2018non} based on the PAC-Bayes framework \cite{mcallester1999some} have made remarkable progress towards explaining how overparametrized neural networks simultaneously achieve low risk and zero empirical risk on benchmark datasets. PAC-Bayes bounds deal with randomized classifiers with posterior and prior distributions $\hat{\rho}$ and $\pi$ respectively. Given that typically one wants to bound the risk of a deterministic classifier $f$ the posterior $\hat{\rho}$ is chosen to be in some sense close to $f$(i.e. it is usually centered at $f$). Then, PAC-Bayes theorems make statements that are roughly of the form
\begin{equation}\label{simple_pac}
\bE\mathcal{L}(\hat{\rho})\leq\bE\hat{\mathcal{L}}(\hat{\rho})+\beta\mathrm{KL}(\hat{\rho}||\pi),
\end{equation}
where $\mathcal{L}(\hat{\rho})$ is the risk, $\hat{\mathcal{L}}(\hat{\rho})$ is the empirical risk and the expectation is over the posterior. The $\beta\mathrm{KL}(\hat{\rho}||\pi)$ term between the prior and posterior acts as a measure of complexity for the classifier.

\begin{figure}[t!]
\centering
\begin{subfigure}{.4\textwidth}
  \centering
  \includegraphics[scale=0.6]{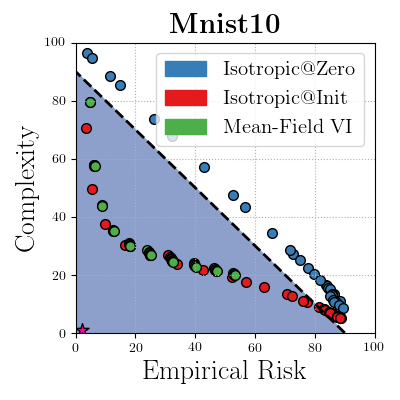}
  \caption{}
  \label{fig1:figure1}
\end{subfigure}%
  \caption{\textbf{Risk-Complexity plot for MNIST 10}: The area below the dashed line corresponds to non-vacuous pairs of (complexity, empirical risk). The purple star corresponds to the optimal bound implied by the testing set. We parametrize the PAC-Bayes bound with different combinations of diagonal Gaussian priors and posteriors. ``Isotropic@Zero" corresponds to isotropic priors and posteriors with the prior centered at 0. ``Isotropic@Init" corresponds to isotropic priors and posteriors with the prior centered at the random initialization. In ``Mean-Field VI" the posterior is diagonal but non-isotropic and we optimize it with Variational Inference. There is negligible improvement over the isotropic case.}
  \label{fig1:figure_full}
\end{figure}

The RHS of \eqref{simple_pac} corresponds to a variational encoding scheme of the deep neural network weights, where the variance of the noise in the posterior measures the level of precision used in the encoding \cite{blier2018description}. In the very influential \citet{dziugaite2017computing}, the authors minimize this variational code directly using a differentiable surrogate, by parameterizing the prior and posterior as Gaussians, and optimizing using stochastic variational inference \cite{hoffman2013stochastic,kingma2013auto}. They obtain non-vacuous generalization bounds on a simplified MNIST\cite{lecun-mnisthandwrittendigit-2010} dataset, but are unable to scale their result to larger problems. 

Stochastic variational inference is know to result in poor weight encodings, but the reasons behind this are unclear \cite{blier2018description}. Variational inference, in the context of Bayesian neural networks, is thought to suffer from high gradient variance \cite{kingma2015variational,wu2018deterministic,wen2018flipout}. In addition, correlations between parameters are often omitted, as storing and manipulating the full covariance matrix is computationally infeasible. This can be seen as adding independent noise to each weight, an approximation know as \emph{mean-field}. This might be too restrictive in deriving useful posteriors \cite{ritter2018scalable,mishkin2018slang}, and therefore tight codes.

Consequently, in \citet{zhou2018non} the authors first compress deep neural networks by sparsifying them and deriving a variational code on the remaining parameters. Off the shelf compression algorithms compress remarkably well and thus \citet{zhou2018non} obtain non-vacuous but loose bounds for the much more complex Imagenet \cite{imagenet_cvpr09}. A significant drawback of this approach is that the bound is derived for a network whose parameters are not similar \emph{even in expectation} to the original ones \cite{suzuki2019compression}.

We thus focus on analyzing the case of applying variational inference directly on the original weights. Importantly, we lack meaningful comparison tools. The techniques in \citet{dziugaite2017computing,zhou2018non} actually provide \emph{multiple} bounds corresponding to different levels of encoding precision of the weights, which is usually controlled by the the parameter $\beta$ in \eqref{simple_pac}. However, results are presented in single (empirical risk, complexity) pairs, making drawing conclusions difficult.

Our first contribution is thus to introduce ``Risk-Complexity" plots \ref{fig1:figure_full}. On the $x$-axis we plot the Empirical Risk $\hat{\mathcal{L}}(\hat{\rho})$, while on the $y$-axis we plot the estimated Complexity $\beta\mathrm{KL}(\hat{\rho}||\pi)$ or the equivalent complexity metric. The plots have a number of advantages. We can easily plot the region of non-vacuity and the location of the best possible bound implied by the \emph{testing} set. For an optimization based bound method we can then derive multiple (complexity, empirical risk) estimates and plot a Pareto front of all combinations. This results in an intuitive way for comparing bounding methods where one can simply inspect the Pareto fronts in relation to the best possible pair implied by the testing set.

Armed with our new visualization tools we are ready to scrutinize the results of \citet{dziugaite2017computing}. The authors combine four elements in deriving non-vacuous bounds: i) changing the prior to be centered at the random initialization instead of at zero ii) optimizing the posterior covariance iii) optimizing the posterior mean iii) simplifying the classification problem by merging the 10 MNIST classes into 2 aggregate ones. In this way it is unclear what is the contribution of each to obtaining non-vacuous bounds. 

In particular, separating the effects of i,ii and iii is important. Flatness at the minimum has been frequently cited as a desirable property for good generalization \cite{keskar2016large}. However, current results show mainly empirical correlations with generalization error \cite{keskar2016large} and the exact effect of flatness is still debated \cite{dinh2017sharp}. Point ii is related to flatness at the minimum, as increased posterior variance while $\bE\hat{\mathcal{L}}(\hat{\rho})$ remains small implies a flat minimum. Importantly, when relating PAC-Bayes to flatness one needs to keep the mean of the posterior fixed. Optimizing the mean and then the covariance corresponds to measuring the flatness of a \emph{different} minimum. 

\textbf{Our contributions.} Through detailed experiments we find that for diagonal Gaussian priors and posteriors the dominant element which turns a vacuous bound to non-vacuous is centering the prior at the random initialization instead of at 0. Optimizing the covariance using stochastic variational inference results in negligible or no gains. In fact, a simple isotropic Gaussian baseline in the prior and posterior results in nearly identical bound values. 

We are then motivated to investigate two common explanations for this ineffectiveness. First it could be that stochastic variational inference has not properly converged. Secondly, PAC-Bayes theory allows improved bounds by choosing priors that reflect prior knowledge about the problem, as long as these priors don't depend on the training set. Choosing the random initialization to be the prior mean is already a good prior mean choice. It might be that through a better choice of prior \emph{covariance} the mean-field approximation could yield meaningful improvements to the posterior covariance and hence the bound.

Through a simple theoretical analysis, we explore both of these explanations. Specifically, we leverage the fact that the loss landscape around the minimum is empirically quadratic, to derive closed form bound solutions with respect to both posterior and prior covariance. The second result is invalid under the PAC-Bayes framework but is useful as a sanity check. Our results imply both problems with optimization of VI as well as that significantly better priors can in theory be found. At the same time, the closed form results are far from optimal and point to intrinsic limitations of the mean-field approximation.

We then motivate modeling the curvature at the minimum through a simplified version of K-FAC \cite{martens2015optimizing}. This allows us to efficiently sample (complexity, empirical risk) pairs with improved curvature estimates. Using our Risk-Complexity plots, we find that for randomized classifiers with medium to low empirical risk this results in significant improvements in the generalization bound quality, compared to the implied limits of the mean-field approximation.

\subsection{Related work}
\textbf{Criticism of uniform convergence.} In \citet{nagarajan2019uniform}, the authors posit that two sided uniform convergence bounds cannot produce non-vacuous estimates for deep neural networks, even with aggressive pruning of the hypothesis space. To the best of the authors understanding the criticism holds only for derandomized PAC-Bayes bounds. In the following we will be dealing only with bounding the generalization error of stochastic classifiers. Even for the deterministic case the issue is far from resolved \cite{negrea2019defense}.

\textbf{Bounds leveraging the Hessian.} A number of bounds incorporating the Hessian have been proposed. Some works provide complexity measures that by design simply correlate with generalization error \cite{keskar2016large,li2019hessian,rangamani2019scale,liang2017fisher,jia2019information}. Others approximate the loss around the minimum using a second order Taylor expansion \cite{tsuzuku2019normalized,wang2018identifying} and and then optimize the bound with respect to this approximation. In \citet{tsuzuku2019normalized} the authors first set the prior variance equal to the posterior variance, and then optimize the bound. This results in a suboptimal choice of prior. In \citet{wang2018identifying} the authors restrict the Hessian to be diagonal and optimize with respect only to the posterior covariance. Both \citet{tsuzuku2019normalized} and \citet{wang2018identifying} result in vacuous bounds.

\textbf{Other bounds and relationship to Bayesian Inference.} There has been a huge number of works on generalization bounds for deep neural networks\cite{bartlett2017spectrally,golowich2017size,wei2019data,ledent2019improved,pitas2019some}. These are typically vacuous by several orders of magnitude. A number of works have pointed out the relationship between PAC-Bayes and Bayesian Inference \cite{germain2016pac,achille2018emergence,achille2019information,dziugaite2017computing}.   

In \citet{huang2019stochastic} the authors propose ``Kronecker flow" to obtain better PAC-Bayes bounds. While we also test a more flexible posterior, our emphasis is on a detailed criticism of the mean-field approximation. Furthermore, as we discuss in section \ref{beyond_mean}, flow based methods face a number of challenges in our testing setup.

\section{Preliminaries}

A neural network transforms it's inputs $\ba_0=\bx$ to an output $f_{\btheta}(\bx)=\ba_l$ through a series of $l$ layers, each of which consists of a bank of units/neurons. The computation performed by each layer $i\in\{1,...,l\}$ is given as follows

\begin{equation*}
\begin{split}
&\bs_i=\bW_i\ba_{i-1},\\
&\ba_i=\phi_i(\bs_{i}),\\
\end{split}
\end{equation*}

where $\phi_i$ is an element-wise non-linear function and $\bW_i$ is a weight matrix.

We will define $\btheta=[\text{vec}(\bW_0)\text{vec}(\bW_0) \cdots \text{vec}(\bW_l)]$, which is the vector consisting of all the network's parameters concatenated together, where $\text{vec}$ is the operator which vectorizes matrices by concatenating their rows horizontally. 

We denote the learning sample $(X,Y)=\{(\bx_i,y_i)\}^n_{i=1}\in(\mathcal{X}\times\mathcal{Y})^n$, that contains $n$ input-output pairs. Samples $(X,Y)$ are assumed to be sampled randomly from a distribution $\mathcal{D}$. Thus, we denote $(X,Y)\sim\mathcal{D}^n$ the i.i.d observation of $n$ elements. We consider loss functions $\ell:\mathcal{F}\times\mathcal{X}\times\mathcal{Y}\rightarrow\mathbb{R}$, where $\mathcal{F}$ is a set of predictors $f:\mathcal{X}\rightarrow\mathcal{Y}$. We also denote the empirical risk $\hat{\mathcal{L}}^{\ell}_{X,Y}(f)=(1/n)\sum_i\ell(f,\bx_i,y_i)$ and the risk $\mathcal{L}^{\ell}_{\mathcal{D}}(f)=\bE_{(\bx,y)\sim\mathcal{D}}\ell(f,\bx,y)$.

We will use two loss functions, the non-differentiable zero one loss $\ell_{01}(f,x,y)=\mathbb{I}(\argmax(f(x))=y)$, and categorical cross-entropy, which is a commonly used differentiable surrogate $\ell_{\text{cat}}(f,x,y)=-\sum_i\mathbb{I}[i=y]\log(f(x)_i)$, where we assume that the outputs of $f$ are normalized to form a probability distribution.

We will also use the following PAC-Bayes formulation, by \citet{catoni2007pac}

\begin{theorem}{\cite{catoni2007pac}}\label{catoni}
Given a distribution $\mathcal{D}$ over $\mathcal{X}\times\mathcal{Y}$, a hypothesis set $\mathcal{F}$, a loss function $\ell':\mathcal{F}\times\mathcal{X}\times\mathcal{Y}\rightarrow[0,1]$, a prior distribution $\pi$ over $\mathcal{F}$, a real number $\delta\in(0,1]$, and a real number $\beta>0$, with probability at least $1-\delta$ over the choice of $(X,Y)\sim\mathcal{D}^n$, we have
\begin{equation}\label{catoni_eq}
\begin{split}
\forall{\hat{\rho}} \;\mathrm{on}\; \mathcal{F}:\bE_{f\sim\hat{\rho}} \mathcal{L}^{\ell'}_{\mathcal{D}}(f) \leq& \Phi^{-1}_\beta(\bE_{f\sim\hat{\rho}}\hat{\mathcal{L}}^{\ell'}_{X,Y}(f)\\
&+\frac{1}{\beta n}(\mathrm{KL}(\hat{\rho}||\pi)+\ln{\frac{1}{\delta}})),\\
\end{split}
\end{equation}
where $\Phi^{-1}_\beta(x) = \frac{1-e^{-\beta x}}{1-e^{-\beta}}$.
\end{theorem}

The above PAC-Bayes theorem works with bounded loss functions and as such is typically evaluated with the zero one loss $\ell_{01}$. However, one might want to optimize the above bound as proposed in \citet{dziugaite2017computing}. One approach, is to then parametrize $f_{\btheta}$ using diagonal Gaussians as $\hat{\rho}(\btheta)=\mathcal{N}(\bmu_{\hat{\rho}},\bsig_{\hat{\rho}})$ and the prior as $\pi(\btheta)=\mathcal{N}(\bmu_{\pi},\lambda\boldsymbol{\mathrm{I}})$. Then, one can use the reparametrization trick $\btheta = \bmu_{\hat{\rho}}+\sqrt{\bsig_{\hat{\rho}}}\odot\mathcal{N}(\boldsymbol{0},\boldsymbol{\mathrm{I}})$ and the categorical cross-entropy to optimize the surrogate

\begin{equation}\label{surrogate}
\bE_{\btheta\sim\hat{\rho}(\btheta)}\hat{\mathcal{L}}^{\ell_{\text{cat}}}_{X,Y}(f_{\btheta})+\frac{1}{\beta n}(\mathrm{KL}(\hat{\rho}(\btheta)||\mathcal{N}(\bmu_{\pi},\lambda\boldsymbol{\mathrm{I}}))+\ln{\frac{1}{\delta}}),
\end{equation}

\begin{figure*}[t!]
\centering
\begin{subfigure}{.3\textwidth}
  \centering
  \includegraphics[scale=0.5]{img/m10.png}
  \caption{}
  \label{fig2:figure1}
\end{subfigure}%
\begin{subfigure}{.3\textwidth}
  \centering
  \includegraphics[scale=0.5]{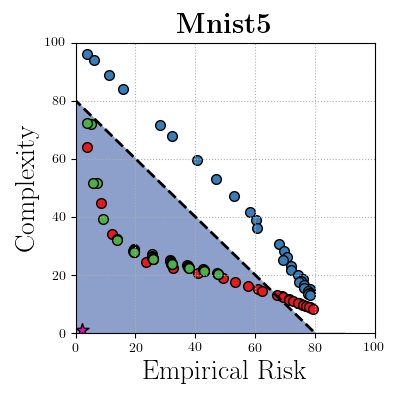}
  \caption{}
  \label{fig2:figure2}
\end{subfigure}%
\begin{subfigure}{.3\textwidth}
  \centering
  \includegraphics[scale=0.5]{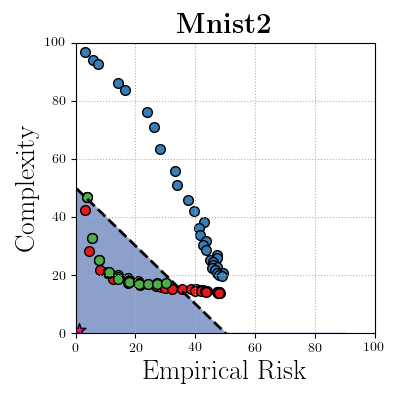}
  \caption{}
  \label{fig2:figure3}
\end{subfigure}%

\begin{subfigure}{.3\textwidth}
  \centering
  \includegraphics[scale=0.5]{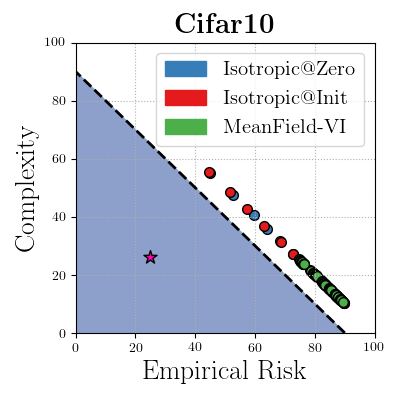}
  \caption{}
  \label{fig2:figure4}
\end{subfigure}%
\begin{subfigure}{.3\textwidth}
  \centering
  \includegraphics[scale=0.5]{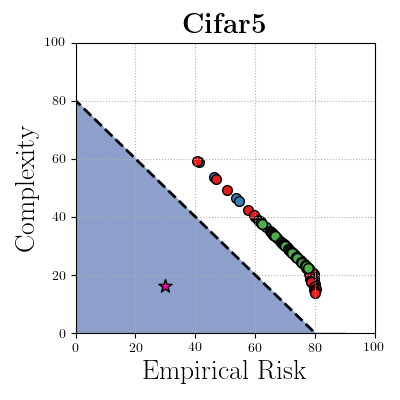}
  \caption{}
  \label{fig2:figure5}
\end{subfigure}%
\begin{subfigure}{.3\textwidth}
  \centering
  \includegraphics[scale=0.5]{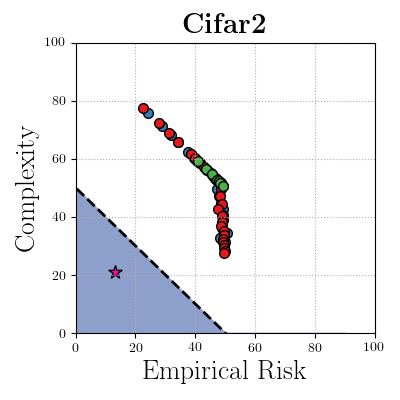}
  \caption{}
  \label{fig2:figure6}
\end{subfigure}%
  \caption{\textbf{Detailed comparison of posterior and prior choices}: The area below the dashed line corresponds to non-vacuous pairs of (complexity, empirical risk). The purple star corresponds to the optimal bound implied by the testing set. For the MNIST case there is a significant improvement when changing from a prior centered at 0 to a prior centered at the random initialization. The baseline isotropic bounds are non-vacuous and optimizing the mean-field approximation using Variational Inference provides no improvements. In the CIFAR case all bounds are vacuous. None of the changes correspond to meaningful improvements in the bound.}
  \label{fig2:figure_full}
\end{figure*}

for $\bmu_{\hat{\rho}}$, $\bsig_{\hat{\rho}}$. In practice, one optimizes \eqref{surrogate}, but wants to evaluate \eqref{catoni_eq}. It's also often beneficial to fine tune $\lambda$ and we want to approximate $\bE_{f\sim\hat{\rho}}\hat{\mathcal{L}}^{\ell_{01}}_{X,Y}(f)$ with an empirical estimate. We take a union bound over values of $\lambda$, and apply a Chernoff bound for the tail of the empirical estimate of$\bE_{f\sim\hat{\rho}}\hat{\mathcal{L}}^{\ell_{01}}_{X,Y}(f)$. Putting everything together, one can obtain valid PAC-Bayes bounds subject to a posterior distribution $\hat{\rho}^*(\btheta)$ that hold with probability at least $1-\delta-\delta'$ and are of the form

\begin{equation}\label{empirical_bound}
\begin{split}
\bE_{\btheta\sim\hat{\rho}^*(\btheta)} \mathcal{L}^{\ell_{01}}_{\mathcal{D}}(f_{\btheta}) \leq& \Phi^{-1}_\beta(\tilde{\mathcal{L}}^{\ell_{01}}_{X,Y}(f_{\btheta})+\frac{1}{\beta n}\mathrm{KL}(\hat{\rho}^*(\btheta)||\pi)\\
&+\frac{1}{\beta n}\ln(\frac{\pi^2b^2\ln(c/\lambda)^2}{6\delta})+\sqrt{\frac{\ln{\frac{2}{\delta'}}}{m}}),\\
\end{split}
\end{equation}
where $\Phi^{-1}_\beta(x) = \frac{1-e^{-\beta x}}{1-e^{-\beta}}$. Also $c,b$ are constants, $m$ is the number of samples from $\hat{\rho}$ for approximating $\bE_{f\sim\hat{\rho}}\hat{\mathcal{L}}^{\ell_{01}}_{X,Y}(f)$ and $\tilde{\mathcal{L}}^{\ell_{01}}_{X,Y}(f_{\btheta})$ the empirical estimate.

It is not difficult to see, that for a high enough number of samples $n$ and $m$, the terms in line 2 of \eqref{empirical_bound} have a negligible effect on the bound. All proofs are deferred to the Appendix.

\section{Empirical results}\label{emp_results}

We tested 6 different datasets. These consist of the original MNIST-10 and CIFAR-10 \cite{krizhevsky2010convolutional} datasets, as well as simplified versions, where we collapsed the 10 classes into 5 and 2 aggregate classes, potentially simplifying the classification problem. All had 50000 training samples. We test the architectures
$$\text{input}\rightarrow300\text{FC}\rightarrow300\text{FC}\rightarrow\#classes\text{FC}\rightarrow\text{output}$$
on MNIST, and 
$$\text{input}\rightarrow200\text{FC}\rightarrow200\text{FC}\rightarrow\#classes\text{FC}\rightarrow\text{output}$$
on CIFAR.

\begin{figure*}[t!]
\centering
\begin{subfigure}{.3\textwidth}
  \centering
  \includegraphics[scale=0.5]{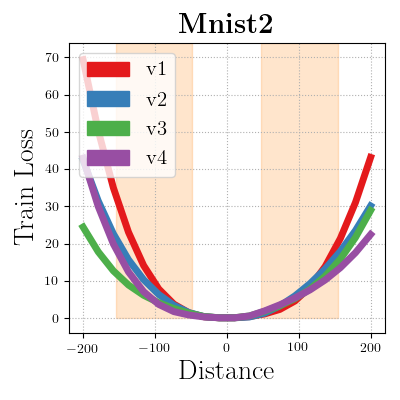}
  \caption{}
  \label{fig3:figure1}
\end{subfigure}%
\begin{subfigure}{.3\textwidth}
  \centering
  \includegraphics[scale=0.5]{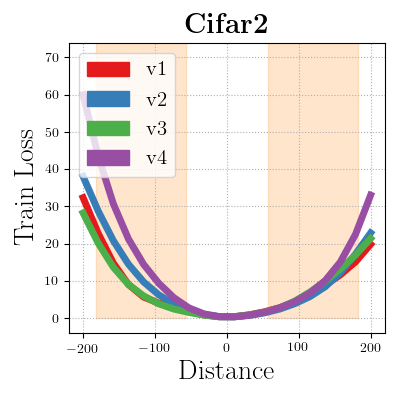}
  \caption{}
  \label{fig3:figure1}
\end{subfigure}%
  \caption{\textbf{Empirical evaluation of the categorical cross-entropy loss}: We take normalized random directions $\bv_i,\; i\in\{1,2,3,4\}$ and plot the deterministic categorical cross-entropy loss $\hat{\mathcal{L}}^{\ell_{\text{cat}}}_{X,Y}(f_{\btheta})$ for MNIST2 and CIFAR2 and values on the line $\btheta=\btheta_*+t \bv_i,\; t\in[-200,200]$. We see that the loss closely reassembles a quadratic our the minimum $\btheta_*$. High dimensional Gaussian vectors concentrate close to a hypersphere centered on the mean. We find the radius of the hyperspheres and shade the corresponding 1 dimensional cross sections in the plots. Posteriors relevant to our experiments concentrate within an area well approximated by the quadratic.}
  \label{fig3:figure_full}
\end{figure*}

We also tested three combinations of prior and posterior 
\begin{enumerate}
\item $\hat{\rho}(\btheta)=\mathcal{N}(\bmu_{\hat{\rho}},\lambda\boldsymbol{\mathrm{I}})$ , $\pi(\btheta)=\mathcal{N}(0,\lambda\boldsymbol{\mathrm{I}})$ 
\item $\hat{\rho}(\btheta)=\mathcal{N}(\bmu_{\hat{\rho}},\lambda\boldsymbol{\mathrm{I}})$ , $\pi(\btheta)=\mathcal{N}(\bmu_{\mathrm{init}},\lambda\boldsymbol{\mathrm{I}})$ 
\item $\hat{\rho}(\btheta)=\mathcal{N}(\bmu_{\hat{\rho}},\bsig_{\hat{\rho}})$ , $\pi(\btheta)=\mathcal{N}(\bmu_{\mathrm{init}},\lambda\boldsymbol{\mathrm{I}})$.  
\end{enumerate}

\textbf{Isotropic posterior.} Isotropic combinations 1 and 2 differ only in the prior mean. The first prior is centered at 0, while the second prior is centered at the random deep neural network initialization. In practice, to derive multiple (complexity, empirical risk) pairs we sample $\lambda$,$\beta$ in the range $\lambda \in [0.031,0.3]$ and $\beta\in[1,5]$. For these we compute $\hat{\mathcal{L}}(\hat{\rho})$ and $\mathrm{KL}(\hat{\rho}||\pi)$. The second can be computed analytically, while we approximate the first using Monte Carlo sampling with $m=1000$ samples from $\hat{\rho}$. We then plug the results into \eqref{empirical_bound}. We set the estimated complexity as $\mathrm{Complexity}\equiv[\Phi^{-1}_{\beta^*}(\hat{\mathcal{L}}(\hat{\rho}^*)+\frac{1}{\beta^* n}\mathrm{KL}(\hat{\rho}^*||\pi))-\hat{\mathcal{L}}(\hat{\rho}^*)]$, where $\beta^*$ is the optimal $\beta$.

\textbf{Diagonal posterior (VI).} The third case corresponds to a non-informative prior centered at the random initialization and a posterior with diagonal covariance. For MNIST we do a grid search over $\beta\in[1,5]$ and $\lambda\in[0.03,0.1]$ while for CIFAR we search in $\beta\in[1,5]$ and $\lambda\in[0.1,0.3]$. For each $(\beta,\lambda)$ pair we optimize $\bsig_{\hat{\rho}}$ using the surrogate \eqref{surrogate}. Specifically, we use the state of the art Flipout estimator \cite{wen2018flipout}. We used 5 epochs of training using the Adam optimizer \cite{kingma2014adam} with a learning rate of $1e-1$. Increasing the number of epochs didn't affect the results. We calculate the complexity and empirical risk as in the isotropic case.

We plot the Pareto fronts of the above samples in \ref{fig2:figure_full}. For the case of MNIST, changing from the prior centered at 0 to the prior centered at the random initialization resulted in a significant improvement of the bound. The resulting bounds with a prior at the random initialization are non-vacuous, even for the simple isotropic posterior. Optimizing the covariance with VI yields negligible or no improvements.  

For CIFAR, we do not see significant variation in the bounds. The Catoni bound has a saturating effect above the line $y=1-x,\; \text{s.t.}\; x\in[0,1]$. All (complexity,empirical risk) pairs fall into this saturating region. Specifically, mean-field VI fails to meaningfully improve the bound. Looking at the optimal bound points (star shapes), one explanation for the difference with MNIST, is that CIFAR DNNs have overfit the data significantly.

\section{Quadratic Approximation}

The stochastic and non-convex objective \eqref{surrogate} is difficult to analyze theoretically. As such we first propose to approximate the cross-entropy loss at the mean of the posterior using a second order Taylor expansion which will make the subsequent analysis tractable. Denoting the centered posterior as $\hat{\rho}'(\btheta)$ we get 
\begin{equation}\label{quad_approx_der}
\begin{split}
&C_{\beta}(X,Y;\hat{\rho},\pi)=\bE_{\btheta\sim\hat{\rho}(\btheta)}\hat{\mathcal{L}}^{\ell_{\text{cat}}}_{X,Y}(f_{\btheta})+\beta\normalfont{\text{KL}}(\hat{\rho}(\btheta)||\pi(\btheta))\\
&\approx \bE_{\bet\sim\hat{\rho}'(\btheta)}[\bet^T \nabla\hat{\mathcal{L}}^{\ell_{\text{cat}}}_{X,Y}(f_{\btheta})+\frac{1}{2}{\bet}^T \nabla^2\hat{\mathcal{L}}^{\ell_{\text{cat}}}_{X,Y}(f_{\btheta}) \bet]\\ 
&+\beta\normalfont{\text{KL}}(\hat{\rho}(\btheta)||\pi(\btheta))\\
&\approx \bE_{\bet\sim\hat{\rho}'(\btheta)}[\frac{1}{2}{\bet}^T \nabla^2\hat{\mathcal{L}}^{\ell_{\text{cat}}}_{X,Y}(f_{\btheta}) \bet]+\beta\normalfont{\text{KL}}(\hat{\rho}(\btheta)||\pi(\btheta)).
\end{split}
\end{equation}
We've made a number of assumptions. First, we assumed that the gradient at the point of expansion is zero. For a well trained overparametrized DNN this is a reasonable assumption. Secondly, we omit terms of the Taylor expansion of order $\geq3$. This results in a quadratic approximation. We conduct experiments to see whether this is reasonable. We take random directions along the loss landscape and plot along them the value of the loss. We see in Figure \ref{fig3:figure_full} that the loss is indeed approximately quadratic around the minimum.  We also note that approximating the loss as quadratic has been used to obtain state of the art results in the DNN compression literature \cite{dong2017learning,wang2019eigendamage,peng2019collaborative,lecun1990optimal,hassibi1993second}. 

For the expectation of the quadratic loss to be a good approximation of the expectation of the categorical loss, the mass of the posterior has to be concentrated at locations where the true loss is well approximated by a quadratic. We have thus far dealt with Gaussian posteriors $\hat{\rho}(\btheta)=\mathcal{N}(\bmu_{\hat{\rho}},\bsig_{\hat{\rho}})$, where $\forall i,\; \sigma_{\hat{\rho}i}\approx\lambda$, $0.01\leq\lambda\leq1$. It is well know that Gaussians in high dimensions concentrate on a thin "bubble" away from the origin. We can make a rough calculation of the radius of this bubble \cite{vershynin2018high}. Specifically, assuming that $\forall i, \; \bsig_{\hat{\rho}i}=\lambda$, we can calculate $ \bE_{\bet\sim\hat{\rho}'(\btheta)}||\bet||_2^2=\bE_{\bet\sim\mathcal{N}(0,\bsig_{\hat{\rho}})}[\sum_{i=0}^{d}\eta_i^2]=\sum_{i=0}^{d}\sigma_{\hat{\rho}i}=\lambda d$. Finally we expect that the radius of the ``bubble" is $ \bE_{\bet\sim\hat{\rho}'(\btheta)}||\bet||_2\approx\sqrt{\lambda d}$. We plot these regions in Figure \ref{fig3:figure_full}. We see that posteriors concentrate within areas where the quadratic approximation is reasonable. 

\subsection{Optimal Posterior}
We make the slightly more general modeling choices $\hat{\rho}(\btheta)=\mathcal{N}(\bmu_{\hat{\rho}},\bSigma_{\hat{\rho}})$ and $\pi(\btheta)=\mathcal{N}(\bmu_{\pi},\lambda\bSigma_{\pi})$. We can then show that the optimal posterior covariance of the objective \eqref{quad_approx_der} for fixed prior and posterior means has a closed form solution.

\begin{lemma}
The convex optimization problem $\min_{\bSigma_{\hat{\rho}} }\bE_{\bet\sim\hat{\rho}'(\btheta)}[\frac{1}{2}\bet^T \bH \bet]+\beta\normalfont{\text{KL}}(\hat{\rho}(\btheta)||\pi(\btheta))$ where $\hat{\rho}(\btheta)=\mathcal{N}(\bmu_{\hat{\rho}},\bSigma_{\hat{\rho}})$ and $\pi(\btheta)=\mathcal{N}(\bmu_{\pi},\lambda\bSigma_{\pi})$ is minimized at
\begin{equation}\label{optimal_posterior_equation}
\bSigma_{{\hat{\rho}}}^* = \beta(\bH+\frac{\beta}{\lambda}\bSigma_{\pi}^{-1})^{-1},
\end{equation}
where $\bH \equiv \nabla^2\hat{\mathcal{L}}^{\ell_{\text{cat}}}_{X,Y}(f_{\btheta})$ captures the curvature at the minimum, while $\bSigma_{\pi}$ is the prior covariance.
\end{lemma}

This can been seen as a \emph{Laplace} approximation \cite{bishop2006pattern} to the posterior. 

\begin{figure*}[t!]
\centering
\begin{subfigure}{.3\textwidth}
  \centering
  \includegraphics[scale=0.5]{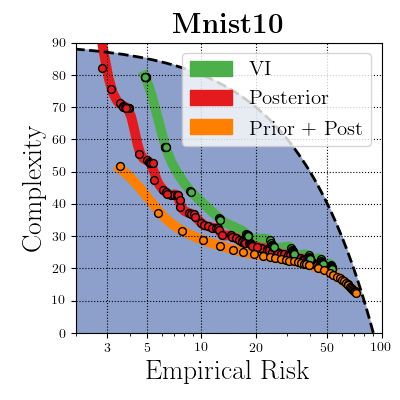}
  \caption{}
  \label{fig4:figure1}
\end{subfigure}%
\begin{subfigure}{.3\textwidth}
  \centering
  \includegraphics[scale=0.5]{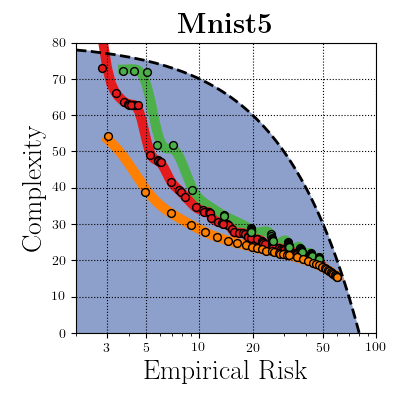}
  \caption{}
  \label{fig4:figure2}
\end{subfigure}%
\begin{subfigure}{.3\textwidth}
  \centering
  \includegraphics[scale=0.5]{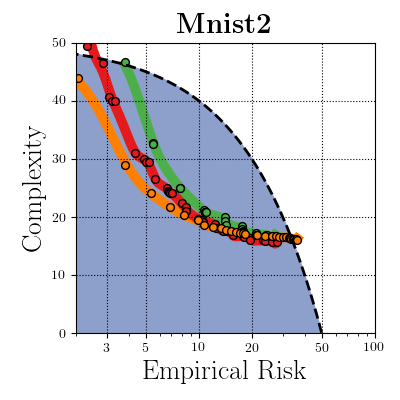}
  \caption{}
  \label{fig4:figure3}
\end{subfigure}%

\begin{subfigure}{.3\textwidth}
  \centering
  \includegraphics[scale=0.5]{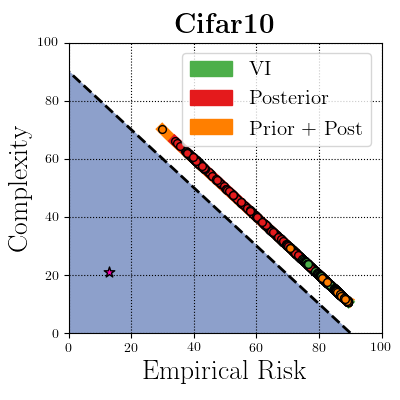}
  \caption{}
  \label{fig4:figure1}
\end{subfigure}%
\begin{subfigure}{.3\textwidth}
  \centering
  \includegraphics[scale=0.5]{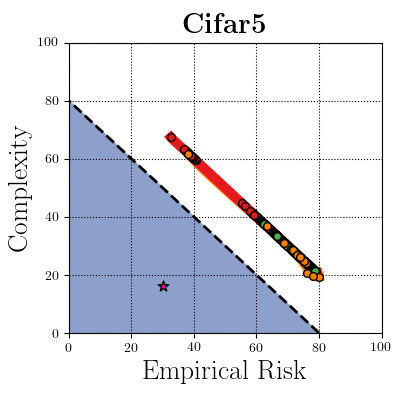}
  \caption{}
  \label{fig4:figure2}
\end{subfigure}%
\begin{subfigure}{.3\textwidth}
  \centering
  \includegraphics[scale=0.5]{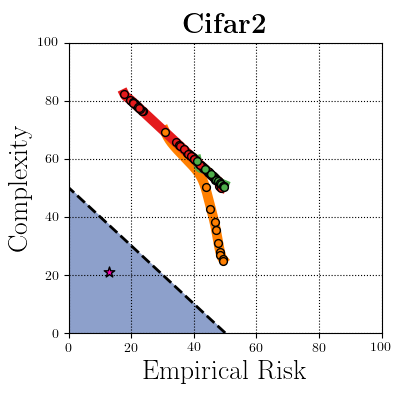}
  \caption{}
  \label{fig4:figure3}
\end{subfigure}%
  \caption{\textbf{Closed form posterior and prior}: We plot the results obtained by mean-field Variational Inference, as well as the closed form bounds with optimized posterior and jointly optimized posterior and prior covariances. For MNIST, we plot the empirical risk in logarithmic scale for ease of exposition. Valid bounds where we only optimize the posterior in closed form get significant benefits over VI of between 5-10\%. Optimizing the prior results in further improvements of 5-10\%, implying that in theory better priors can be found. The results are far from tight even when optimizing the prior and for CIFAR all bounds are vacuous. This implies inherent limitations of the mean-field approximation, as we typically don't even have access to the optimal prior covariance.}
  \label{fig4:figure_full}
\end{figure*}

\subsection{Optimal Prior}

PAC-Bayesian theory allows one to choose an informative prior, however the prior can only depend on the data generating distribution and \emph{not} the training set. A number of previous works \cite{parrado2012pac,catoni2003pac,ambroladze2007tighter} have used this insight mainly on simpler linear settings and usually by training a classifier on a separate training set and using the result as a prior. Recently, \citet{dziugaite2018data} have proposed to use the original training set to derive valid priors by imposing differential privacy constraints.

We ignore these concerns for the moment, and optimize the prior covariance directly. The objective is non-convex, however for the case of \emph{diagonal} prior and posterior covariances we can find the global minimum.

\begin{lemma}\label{thm_opt_prior}
The optimal prior and posterior covariances for $\min_{\bsig_{\hat{\rho}},\bsig_{\pi}}\bE_{\bet\sim\hat{\rho}'(\btheta)}[\frac{1}{2}\bet^T \bH\bet]+\beta\normalfont{\text{KL}}(\hat{\rho}(\btheta)||\pi(\btheta))$ with $\hat{\rho}(\btheta)=\mathcal{N}(\bmu_{\hat{\rho}},\bsig_{\hat{\rho}})$ and $\pi(\btheta)=\mathcal{N}(\bmu_{\pi},\lambda\bsig_{\pi})$ 
have elements

\begin{equation}\label{opt_posterior_diag}
(\sigma_{\hat{\rho}i}^*)^{-1}=\frac{1}{2\beta}[h_{i}+\sqrt{h_{i}^2+\frac{4\beta h_{i}}{(\mu_{i\hat{\rho}}-\mu_{i\pi})^2}}],
\end{equation}

\begin{equation}\label{opt_prior_diag}
(\sigma_{\pi i}^*)^{-1}=\frac{\lambda}{2\beta}[\sqrt{h_{i}^2+\frac{4\beta h_{i}}{(\mu_{i\hat{\rho}}-\mu_{i\pi})^2}}-h_{i}],
\end{equation}
where $\bH \equiv \nabla^2\hat{\mathcal{L}}^{\ell_{\text{cat}}}_{X,Y}(f_{\btheta})$ captures the curvature at the minimum.
\end{lemma}

We \emph{cannot} prove generalization using this result. Rather we use it as a sanity check for what is achievable through the mean-field approximation and an optimal informative prior covariance. 

To approximate the Hessian we note that for the cross entropy loss and the softmax activation function $p(y=c|f_{\btheta})=\mathrm{exp}(f_{\btheta}(\bx)_c)/\sum_i\mathrm{exp}(f_{\btheta}(\bx)_i)$ the Fisher Information matrix coincides with the generalized Gauss-Newton approximation of the Hessian \cite{kunstner2019limitations}. We sample one ouput $\tilde{y}_i$ from the model distribution $p(y_i|f_{\btheta}(\bx_i))$ for each input $\bx_i$, and approximate $\bH\approx \sum_{i=0}^n\nabla_{\btheta}\log p(\tilde{y}_i|f_{\btheta}(\bx_i))\nabla_{\btheta}\log p(\tilde{y}_i|f_{\btheta}(\bx_i))^{\mathrm{T}}$, retaining only the diagonal elements.

Keeping the posterior and prior means fixed, we optimize the posterior covariance, as well as the posterior and \emph{prior} covariance jointly in closed form. We plot the results in Figure \ref{fig4:figure_full}, using the same approach as section \ref{emp_results} with $m=1000$ for \eqref{opt_posterior_diag},\eqref{opt_prior_diag}, $m=100$ for \eqref{optimal_posterior_equation} and sampling over $\beta$ and $\lambda$. For MNIST, valid bounds where we only optimize the posterior in closed form get significant benefits over VI of between 5-10\%. Thus, even though Adam is very robust to hyperparameter selection, and the Flipout estimator is state of the art, one might look to hyperparameter tuning for better results. We present arguments in the next section, that hold also for the mean-field case, as to why it should be beneficial to avoid hyperparameter tuning. Invalid bounds where we optimize the prior and posterior jointly result in further improvements of 5-10\%, implying that in theory better priors can be found. The bounds are far from tight, even when optimizing the prior, and for CIFAR all bounds are vacuous. This implies that the mean-field approximation is limited in the bound improvements it can provide.

\section{Beyond the mean-field approximation}\label{beyond_mean}

\subsection{Computational Issues}
A number of approximations exist to model richer posteriors. In \citet{mishkin2018slang}, the authors model the covariance as having a low-rank + diagonal structure. In normalizing flows \citet{rezende2015variational} a simple initial density is transformed into a more complex one, by applying a sequence of invertible transformations, until a desired level of complexity is attained. In K-FAC \citet{martens2015optimizing}, the Hessian can be approximated as a Khatri-Rao product to construct a Laplace approximation of the posterior \cite{ritter2018scalable}. 

\begin{figure*}[t!]
\centering
\begin{subfigure}{.3\textwidth}
  \centering
  \includegraphics[scale=0.5]{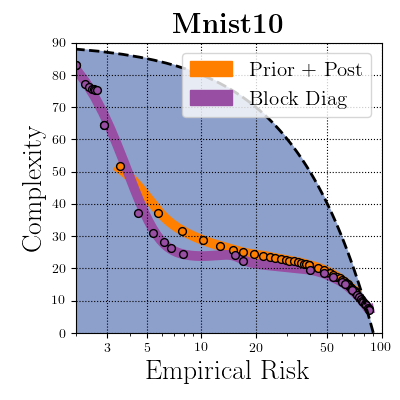}
  \caption{}
  \label{fig5:figure1}
\end{subfigure}%
\begin{subfigure}{.3\textwidth}
  \centering
  \includegraphics[scale=0.5]{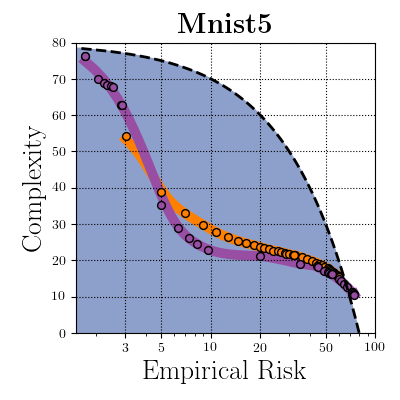}
  \caption{}
  \label{fig5:figure2}
\end{subfigure}%
\begin{subfigure}{.3\textwidth}
  \centering
  \includegraphics[scale=0.5]{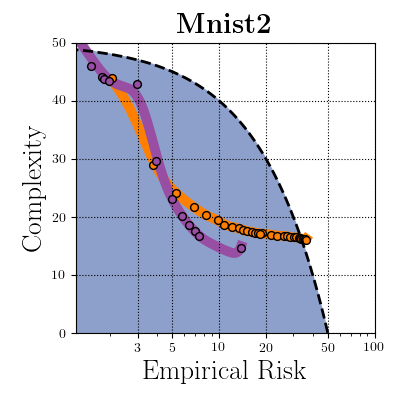}
  \caption{}
  \label{fig5:figure3}
\end{subfigure}%
  \caption{\textbf{Beyond the mean field approximation}: We compared the simplified K-FAC curvature approximation to the closed form \emph{invalid} mean-field inference. Invalid results correspond to an optimal prior and posterior covariance to which we don't typically have access. For medium to low empirical risk the block diagonal curvature improves the bound for MNIST10-5-2 by $8.2\%, 7.5\%, 4.4\%$ respectively. }
  \label{fig5:figure_full}
\end{figure*}

\textbf{Optimizing multiple variational objectives.} To obtain Pareto fronts we will perform a grid search over $\lambda$ and $\beta$, corresponding to $\mathcal{O}(10^2)$ classifiers with different empirical risk and complexity. Optimizing variational objectives is known to be unstable, to scale badly and to require extensive hyperparameter tuning \cite{wu2018deterministic}. Optimizing each posterior using SGD as in \citet{mishkin2018slang,rezende2015variational}, for even a few minutes, can add several hours to obtaining the full grid. Hyperparameter tuning objectives that do not converge can quickly make the task infeasible.

\textbf{Sampling efficiently from the posterior.} At the same time we will need to sample efficiently between $\mathcal{O}(10^4)$ and $\mathcal{O}(10^5)$ posterior samples. This is because we will be applying a Chernoff bound on the tail of the empirical risk. For flow based methods, the KL term also has to be approximated with MC sampling. 

In the non-flow based methods, one typically seeks to factor $\Sigma = LL^T$. Then $y=Lz$, where z is standard normal, has the appropriate distribution, and can be sampled efficiently. While \citet{mishkin2018slang} provide an efficient Cholesky factorization of their low-rank + diagonal approximation, the Khatri-Rao product \cite{martens2015optimizing} has no obvious Cholesky factorization. Inference time in flow based methods will be influenced by the number of mappings used in the flow.

\textbf{Simplified K-FAC Laplace.} 
We assume a multiclass classification problem with $c$ classes, and that the labels $y$ are one-hot encoded. We then define the mean square error loss $\ell_{\text{mse}}(f,x,y)=(1/c)\sum_{i=0}^c(f(x)_i-y_i)^2$. Assuming $r$ neurons per layer, $\btheta$ has a form $\btheta=[\text{vec}(\bW_0^{0,:})\text{vec}(\bW_0^{1,:}) \cdots \text{vec}(\bW_l^{r,:})]$. We also denote for layer $i$ and neuron $j$, $\btheta_{ij}$, $\bmu_{\hat{\rho}ij}$, $\bSigma_{\hat{\rho}ij}$, $\bmu_{\pi ij}$ the corresponding split variables. We can then motivate optimizing the following surrogate upper bound 
\begin{lemma}
Assuming negligible layerwise derivatives of order other than 2, the differentiable surrogate objective
\begin{equation}\label{surrogate_mse}
\bE_{\btheta\sim\hat{\rho}(\btheta)}\hat{\mathcal{L}}^{\ell_{\text{mse}}}_{X,Y}(f_{\btheta})+\frac{1}{\beta n}(\mathrm{KL}(\hat{\rho}(\btheta)||\mathcal{N}(\bmu_{\pi},\lambda\boldsymbol{\mathrm{I}}))+\ln{\frac{1}{\delta}}),
\end{equation}
has the following upper bound
\begin{equation}\label{surrogate_mse_upper}
\begin{split}
&\sum_{i,j} [\bE_{\bet_{ij} \sim \hat{\rho}_{ij}'(\btheta)}[\frac{1}{2}\bet_{ij}^T \bH_{i} \bet_{ij}] +\frac{1}{\beta n}\mathrm{KL}(\hat{\rho}_{ij}(\btheta)||\pi_{ij}(\btheta)]\\
&+\mathcal{O}(c^l),\\
\end{split}
\end{equation}
where $\hat{\rho}_{ij}(\btheta)=\mathcal{N}(\bmu_{\hat{\rho}ij},\bSigma_{\hat{\rho}ij})$, $\pi_{ij}(\btheta)=\mathcal{N}(\bmu_{\pi ij},\lambda\boldsymbol{\mathrm{I}})$, $\bH_{i}=(1/n)\sum_{k=0}^n \ba_i^k{\ba_i^k}^T$, are neuronwise posteriors, priors and Hessians.
\end{lemma}
The above corresponds to a greatly simplified version of K-FAC, where each layer has a posterior with covariance $\bSigma_i=\bH_i\otimes\boldsymbol{\mathrm{I}}$, i.e. we assume correlations only for parameters in each neuron. While this approximation covers our needs, one could in principle use the slightly more expressive $\bSigma_i=\bH_i\otimes\bG_i$, where $\bG_i = \bE[\bg_i \bg_i^T]$ and $\bg_i$ are the backpropagated layerwise errors for layer $i$ \cite{ritter2018scalable}. We've broken the original into many much smaller subproblems. We can compute the Hessian efficiently and in a stable way once, and then sample the posterior efficiently in closed form at different variance levels $\lambda$.

\subsection{Empirical Results}
We now present results on the MNIST datasets. We run a grid search over $\beta$ and $\lambda$, with 20 samples each, for $\beta \in [0.001,0.02]$ and $\lambda \in [0.001,0.1]$. We use $m=1000$ samples for estimating the empirical risk. For computing the Pareto fronts we optimize \eqref{surrogate_mse_upper} with \eqref{optimal_posterior_equation} and evaluate \eqref{empirical_bound} following the procedure of Section \ref{emp_results}. The running time for each experiment was 33h, 30h and 25h respectively.

We plot the results in Figure \ref{fig4:figure_full}. We compare with the invalid priors and posteriors which we cannot typically access. At very low and very high empirical risk levels the complexity estimates saturate. However, for medium empirical risk levels the block diagonal covariance yields significant improvements to the bounds. The effect is more pronounced on the more difficult MNIST10 and MNIST5 experiments, where using the block diagonal posterior results in a decrease in the estimated complexity of $\sim10\%$.

\section{Conclusion and Future Work}
We have presented several arguments in favor of richer posterior distributions under the PAC-Bayes framework. We've only scratched the surface, as we've relaxed only slightly from the diagonal case, getting significant gains. Of course, another line of approach would be to optimize further the prior \emph{mean} in a valid way, an area that has been little investigated. As research moves closer to solving the generalization puzzle of deep learning, we hope that our plots provide a more intuitive way to compare new bounds.

\bibliography{icml_sub}
\bibliographystyle{icml2020}

\onecolumn

\section*{Appendix}
\section*{A. Derivations for valid bound}
We present again for clarity the PAC-Bayes bound by \citet{catoni2007pac}.
\begin{customthm}{2.1}{\cite{catoni2007pac}}\label{catoni_appendix}
Given a distribution $\mathcal{D}$ over $\mathcal{X}\times\mathcal{Y}$, a hypothesis set $\mathcal{F}$, a loss function $\ell':\mathcal{F}\times\mathcal{X}\times\mathcal{Y}\rightarrow[0,1]$, a prior distribution $\pi$ over $\mathcal{F}$, a real number $\delta\in(0,1]$, and a real number $\beta>0$, with probability at least $1-\delta$ over the choice of $(X,Y)\sim\mathcal{D}^n$, we have
\begin{equation}\label{catoni_eq_appendix}
\begin{split}
\forall{\hat{\rho}} \;\mathrm{on}\; \mathcal{F}:\bE_{f\sim\hat{\rho}} \mathcal{L}^{\ell'}_{\mathcal{D}}(f) \leq& \Phi^{-1}_\beta(\bE_{f\sim\hat{\rho}}\hat{\mathcal{L}}^{\ell'}_{X,Y}(f)\\
&+\frac{1}{\beta n}(\mathrm{KL}(\hat{\rho}||\pi)+\ln{\frac{1}{\delta}})),\\
\end{split}
\end{equation}
where $\Phi^{-1}_\beta(x) = \frac{1-e^{-\beta x}}{1-e^{-\beta}}$.
\end{customthm}

Evaluating a valid PAC-Bayes bound, using empirical estimates, requires some care. 

\textbf{Optimizing $\lambda$.} For a start, when modeling $\pi(\btheta)=\mathcal{N}(0,\lambda\bI)$, it is often beneficial to optimize the hyperparameter $\lambda$. As the PAC-Bayes theorem requires the prior to be independent from the posterior, we need to take a union bound over an appropriately chosen grid, representing different possible values of $\lambda$. Following \cite{dziugaite2017computing}, we can choose $\lambda=c\exp\{-j/b\}$ for $j\in\mathbb{N}$ and fixed $b,c\geq0$, where $c$ corresponds to the grid scale and $b$ to it's precision. Then, if the PAC-Bayes bound for each $j\in\mathbb{N}$ is designed to hold with probability at least $1-\frac{6\delta}{\pi^2j^2}$, by union bound it will hold uniformly for all $j\in\mathbb{N}$ with probability at least $1-(\frac{6\delta}{\pi^2})\sum_{j\in\mathbb{N}}\frac{1}{j^2} =1-\delta$. We solve for $j=b\log{\frac{c}{\lambda}}$ and substitute this value in the probability for each term in the union bound. We get that \emph{any} bound corresponding to $j\in\mathbb{N}$ holds with probability $1-\frac{6\delta}{\pi^2b^2\ln{(c/\lambda^2)}}$. Thus looking back to theorem \ref{catoni_appendix} the term $\ln{\frac{1}{\delta}}$ becomes $\ln{\frac{\pi^2b^2\ln{(c/\lambda^2)}}{6\delta}}$. In practice we see that even for very large numbers $c,b,\delta$ when divided by the number of samples $n$ the term $\ln{\frac{\pi^2b^2\ln{(c/\lambda^2)}}{6\delta}}$ is negligible and we treat $j$ as a continuous number.

\textbf{Empirical estimate of $\bE_{\btheta\sim\hat{\rho}^*(\btheta)}\hat{\mathcal{L}}^{\ell'}_{X,Y}(f_{\btheta})$.}  Furthermore, assuming an optimized posterior $\hat{\rho}^*(\btheta)$ directly evaluating $\bE_{\btheta\sim\hat{\rho}^*(\btheta)}\hat{\mathcal{L}}^{\ell'}_{X,Y}(f_{\btheta})$ is intractable. Instead, since $\hat{\mathcal{L}}^{\ell'}_{X,Y}(f_{\btheta})$ is a bounded random variable, one can approximate the expectation using Monte Carlo sampling and use a Chernoff bound to bound it's tail. Let $\tilde{\mathcal{L}}^{\ell'}_{X,Y}(f_{\btheta})\equiv (1/m)\sum_{i=0}^{m}\hat{\mathcal{L}}^{\ell'}_{X,Y}(f_{\btheta_i})$ be the observed failure rate of $m$ random hypotheses drawn according to $\hat{\rho}^*(\btheta)$. One can then show the following \cite{langford2002not} (presented here without proof)

\begin{theorem}(Sample Convergence Bound) 
For all distributions, $\hat{\rho}^*(\btheta)$, for all sample sets $(X,Y)$, assuming that $\hat{\mathcal{L}}^{\ell'}_{X,Y}(f_{\btheta}) \in [0,1]$
\begin{equation}
\begin{split}
&\mathrm{Pr}_{\hat{\rho}^*(\btheta)}(\bE_{\btheta\sim\hat{\rho}^*(\btheta)}\hat{\mathcal{L}}^{\ell'}_{X,Y}(f_{\btheta})\leq\tilde{\mathcal{L}}^{\ell'}_{X,Y}(f_{\btheta})+\sqrt{\frac{\ln{\frac{2}{\delta'}}}{m}})\\
&\leq\delta',\\
\end{split}
\end{equation}
where $m$ is the number of evaluations of the stochastic hypothesis.
\end{theorem}

We take a union bound over values of $\lambda$, and apply the Chernoff bound for the tail of the empirical estimate of $\bE_{f\sim\hat{\rho}}\hat{\mathcal{L}}^{\ell'}_{X,Y}(f)$. Putting everything together, one can obtain valid PAC-Bayes bounds subject to a posterior distribution $\hat{\rho}^*(\btheta)$ that hold with probability at least $1-\delta-\delta'$ and are of the form

\begin{equation}\label{empirical_bound_appendix}
\begin{split}
\bE_{\btheta\sim\hat{\rho}^*(\btheta)} \mathcal{L}^{\ell'}_{\mathcal{D}}(f_{\btheta}) \leq& \Phi^{-1}_\beta(\tilde{\mathcal{L}}^{\ell'}_{X,Y}(f_{\btheta})+\frac{1}{\beta n}\mathrm{KL}(\hat{\rho}^*(\btheta)||\pi)\\
&+\frac{1}{\beta n}\ln(\frac{\pi^2b^2\ln(c/\lambda)^2}{6\delta})+\sqrt{\frac{\ln{\frac{2}{\delta'}}}{m}}),\\
\end{split}
\end{equation}
where $\Phi^{-1}_\beta(x) = \frac{1-e^{-\beta x}}{1-e^{-\beta}}$. Also $c,b$ are constants, $m$ is the number of samples from $\hat{\rho}$ for approximating $\bE_{f\sim\hat{\rho}}\hat{\mathcal{L}}^{\ell'}_{X,Y}(f)$ and $\tilde{\mathcal{L}}^{\ell'}_{X,Y}(f_{\btheta})$ the empirical estimate. 

\textbf{Number of samples for Chernoff bound.} In our experiments we use $m=1000$ for all experiments including VI experiments. We make a single exception due to time constraints for the case of optimizing the posterior in closed form (Section 4.1 equation 6) where we use $m=100$.

\begin{itemize}
\item For $m=1000$ and $\delta'=0.05$, this gives bounds with confidence $\sqrt{\frac{\log{(2/0.05)}}{1000}}\approx0.06$. 
\item For $m=100$ and $\delta'=0.05$, this gives bounds with confidence $\sqrt{\frac{\log{(2/0.05)}}{100}}\approx0.19$. 
\end{itemize}

\textbf{Importantly} bounds with even higher confidence $\sqrt{\frac{\log{(2/0.05)}}{10000}}\approx0.019$ and sample size $m=\mathcal{O}(10^4)$ are possible for all experiments with a computational time in the order of weeks. However we consider this point a technicality as the Chernoff bound is quite pessimistic. Empirically the estimates in our experiments converge much faster than implied by the bound analysis, exhibiting no significant difference between $m=1000$, $m=100$ or even $m=10$ in the isotropic cases. This is because this particular Chernoff bound is an application of Hoeffding's inequality for general bounded random variables \cite{vershynin2018high}[p. 25]. The only assumption is that the random variable is bounded $\hat{\mathcal{L}}^{\ell'}_{X,Y}(f_{\btheta}) \in [0,1]$
, and thus the variance of the random variable is significantly overestimated.

\clearpage

\section*{B. Proof of Lemma 4.1}

\begin{customlemma}{4.1}
The convex optimization problem $\min_{\bSigma_{\hat{\rho}}}\bE_{\bet\sim\hat{\rho}'(\btheta)}[\frac{1}{2}\bet^T \bH \bet]+\beta\normalfont{\text{KL}}(\hat{\rho}(\btheta)||\pi(\btheta))$ where $\hat{\rho}(\btheta)=\mathcal{N}(\bmu_{\hat{\rho}},\bSigma_{\hat{\rho}})$ and $\pi(\btheta)=\mathcal{N}(\bmu_{\pi},\lambda\bSigma_{\pi})$ is minimized at
\begin{equation}\label{optimal_posterior_equation_ap}
\bSigma_{\hat{\rho}}^* = \beta(\bH+\frac{\beta}{\lambda}\bSigma_{\pi}^{-1})^{-1},
\end{equation}
where $\bH \equiv \nabla^2\hat{\mathcal{L}}^{\ell_{\text{cat}}}_{X,Y}(f_{\btheta})$ captures the curvature at the minimum, while $\bSigma_{\pi}$ is the prior covariance.
\end{customlemma}
\begin{proof}
\begin{equation}\label{developed_obj}
\begin{split}
C_{\beta}(X,Y;\hat{\rho},\pi)&=\bE_{\bet\sim\hat{\rho}'(\btheta)}[\frac{1}{2}\bet^T \bH \bet]+\beta\normalfont{\text{KL}}(\hat{\rho}(\btheta)||\pi(\btheta))\\
&=\bE_{\bet\sim\hat{\rho}'(\btheta)}[\frac{1}{2} \trace(\bH \bet \bet^T)]+\beta\normalfont{\text{KL}}(\hat{\rho}(\btheta)||\pi(\btheta))\\
&=\frac{1}{2} \trace(\bH\bE_{\bet\sim\hat{\rho}'(\btheta)}[ \bet \bet^T])+\beta\normalfont{\text{KL}}(\hat{\rho}(\btheta)||\pi(\btheta))\\
&=\frac{1}{2} \trace(\bH\bSigma_{\hat{\rho}})+\frac{\beta}{2}(\trace(\frac{1}{\lambda}\bSigma_{\pi}^{-1}\bSigma_{\hat{\rho}})-k+\frac{1}{\lambda}(\bmu_{\hat{\rho}}-\bmu_{\pi})^{\mathrm{T}}\bSigma_{\pi}^{-1}(\bmu_{\hat{\rho}}-\bmu_{\pi})\\
&+\ln\left(\frac{\det\lambda\bSigma_{\pi}}{\det\bSigma_{\hat{\rho}}}\right))\\
\end{split}
\end{equation}

The gradient with respect to $\bSigma_{\hat{\rho}}$ is

\begin{equation}
\frac{\partial C_{\beta}(X,Y;\hat{\rho},\pi)}{\partial \bSigma_{\hat{\rho}}} = [\frac{1}{2}\bH+\frac{\beta}{2\lambda}\bSigma_{\pi}^{-1}-\frac{\beta}{2}\bSigma_{\hat{\rho}}^{-1}].
\end{equation}

Setting it to zero, we obtain the minimizer $\bSigma_{\hat{\rho}}^* = \beta(\bH+\frac{\beta}{\lambda}\bSigma_{\pi}^{-1})^{-1}$. 
\end{proof}
\
\section*{C. Proof of Lemma 4.2}

\begin{customlemma}{4.2}
The optimal prior and posterior covariances for $\min_{\bsig_{\hat{\rho}},\bsig_{\pi}} C_{\beta}(X,Y;\hat{\rho},\pi)=\min_{\bsig_{\hat{\rho}},\bsig_{\pi}}\bE_{\bet\sim\hat{\rho}'(\btheta)}[\frac{1}{2}\bet^T \bH\bet]+\beta\normalfont{\text{KL}}(\hat{\rho}(\btheta)||\pi(\btheta))$ with $\hat{\rho}(\btheta)=\mathcal{N}(\bmu_{\hat{\rho}},\bsig_{\hat{\rho}})$ and $\pi(\btheta)=\mathcal{N}(\bmu_{\pi},\lambda\bsig_{\pi})$ have elements

\begin{equation}
(\sigma_{\hat{\rho}i}^*)^{-1}=\frac{1}{2\beta}[h_{i}+\sqrt{h_{i}^2+\frac{4\beta h_{i}}{(\mu_{i\hat{\rho}}-\mu_{i\pi})^2}}],
\end{equation}

\begin{equation}
(\sigma_{\pi i}^*)^{-1}=\frac{\lambda}{2\beta}[\sqrt{h_{i}^2+\frac{4\beta h_{i}}{(\mu_{i\hat{\rho}}-\mu_{i\pi})^2}}-h_{i}],
\end{equation}

where $\bH \equiv \nabla^2\hat{\mathcal{L}}^{\ell_{\text{cat}}}_{X,Y}(f_{\btheta})$ captures the curvature at the the minimum. Then

\begin{equation}
\begin{split}
\min_{\bsig_{\hat{\rho}},\bsig_{\pi}} C_{\beta}(X,Y;\hat{\rho},\pi)&\geq\frac{1}{2}(\sum_i a_{i}(\mu_{i\hat{\rho}}-\mu_{i\pi})^2\\
&+\beta\sum_i\ln(\frac{h_{i}+a_{i}}{a_{i}})),\\
\end{split}
\end{equation}

where $a_{i} \triangleq a_{i}(\beta,\mu_{i\hat{\rho}},\mu_{i\pi},h_{i}) = \frac{1}{2}[\sqrt{h_{i}^2+\frac{4\beta h_{i}}{(\mu_{i\hat{\rho}}-\mu_{i\pi})^2}}-h_{i}]$.
\end{customlemma}

\begin{proof}
The developed objective \eqref{developed_obj} is 

\begin{equation}\label{developed5}
C_{\beta}(X,Y;\hat{\rho},\pi)=\frac{1}{2} \trace(\bH\bSigma_{\hat{\rho}})+\frac{\beta}{2}(\trace(\frac{1}{\lambda}\bSigma_{\pi}^{-1}\bSigma_{\hat{\rho}})-k+\frac{1}{\lambda}(\bmu_{\hat{\rho}}-\bmu_{\pi})^{\mathrm{T}}\bSigma_{\pi}^{-1}(\bmu_{\hat{\rho}}-\bmu_{\pi})+\ln\left(\frac{\det\lambda\bSigma_{\pi}}{\det\bSigma_{\hat{\rho}}}\right))
\end{equation}

We substitute the precision matrix $\bLambda_{\pi} = \bSigma_{\pi}^{-1}$ and $\bSigma_{\hat{\rho}}$ with the minimizer $\bSigma_{\hat{\rho}}^* = \beta(\bH+\frac{\beta}{\lambda}\bLambda_{\pi})^{-1}$ in \eqref{developed5}, we obtain

\begin{equation}\label{solved_prior_ap}
\begin{split}
C_{\beta}(X,Y;\hat{\rho},\pi)|_{\bSigma_{\hat{\rho}}=\bSigma_{\hat{\rho}}^*}&=\frac{1}{2} \trace(\bH\beta(\bH+\frac{\beta}{\lambda}\bLambda_{\pi})^{-1})+\frac{\beta}{2}(\trace(\frac{1}{\lambda}\bLambda_{\pi}\beta(\bH+\frac{\beta}{\lambda}\bLambda_{\pi})^{-1})\\
&+\frac{1}{\lambda}(\bmu_{\hat{\rho}}-\bmu_{\pi})^{\mathrm{T}}\bLambda_{\pi}(\bmu_{\hat{\rho}}-\bmu_{\pi})-k+\ln\left(\frac{\det\lambda\bLambda_{\pi}^{-1}}{\det\beta(\bH+\frac{\beta}{\lambda}\bLambda_{\pi})^{-1}}\right))\\
=&\frac{\beta}{2} \trace(\bH(\bH+\frac{\beta}{\lambda}\bLambda_{\pi})^{-1})+\frac{\beta^2}{2\lambda}(\trace(\bLambda_{\pi}(\bH+\frac{\beta}{\lambda}\bLambda_{\pi})^{-1}))\\
&+\frac{\beta}{2}(+\frac{1}{\lambda}(\bmu_{\hat{\rho}}-\bmu_{\pi})^{\mathrm{T}}\bLambda_{\pi}(\bmu_{\hat{\rho}}-\bmu_{\pi})-k+\ln\left(\frac{\det\lambda\bLambda_{\pi}^{-1}}{\det\beta(\bH+\frac{\beta}{\lambda}\bLambda_{\pi})^{-1}}\right))\\
=&\frac{\beta}{2}(\trace((\bH+\frac{\beta}{\lambda}\bLambda_{\pi})(\bH+\frac{\beta}{\lambda}\bLambda_{\pi})^{-1})\\
&\frac{1}{\lambda}(\bmu_{\hat{\rho}}-\bmu_{\pi})^{\mathrm{T}}\bLambda_{\pi}(\bmu_{\hat{\rho}}-\bmu_{\pi})-k+\ln\left(\frac{\det\lambda\bLambda_{\pi}^{-1}}{\det\beta(\bH+\frac{\beta}{\lambda}\bLambda_{\pi})^{-1}}\right))\\
=&\frac{\beta}{2}[+\frac{1}{\lambda}(\bmu_{\hat{\rho}}-\bmu_{\pi})^{\mathrm{T}}\bLambda_{\pi}(\bmu_{\hat{\rho}}-\bmu_{\pi})+\ln\left(\frac{\det\lambda\bLambda_{\pi}^{-1}}{\det\beta(\bH+\frac{\beta}{\lambda}\bLambda_{\pi})^{-1}}\right)].\\
\end{split}
\end{equation}

Substituting $\bLambda_{\pi} = \mathrm{diag}(\Lambda_{1{\pi}},\Lambda_{2{\pi}},...,\Lambda_{k{\pi}})$ and $\bH = \mathrm{diag}(h_{1},h_{2},...,h_{k})$ in the above expression we get

\begin{equation}
C_{\beta}(X,Y;\hat{\rho},\pi)|_{\bSigma_{\hat{\rho}}=\bSigma_{\hat{\rho}}^*}=\frac{\beta}{2}(\frac{1}{\lambda}\sum_i\Lambda_{i{\pi}}(\mu_{i{\hat{\rho}}}-\mu_{i\pi})^2-\sum_i\ln(\frac{\Lambda_{i{\pi}}}{\lambda})+\sum_i\ln(\frac{h_{i}+\frac{\beta}{\lambda}\Lambda_{i{\pi}}}{\beta}))
\end{equation}

The above expression is easy to optimize. We see that the sole stationary point exists at

\begin{equation}
\Lambda_{i{\pi}}^*=\frac{\lambda}{2\beta}[\sqrt{h_{i}^2+\frac{4\beta h_{i}}{(\mu_{i{\hat{\rho}}}-\mu_{i\pi})^2}}-h_{i}].
\end{equation}

We now need to calculate second derivatives so as to prove that the stationary point is a local optimum. We go back to the developed objective \eqref{developed5}, and substitute $\bSigma_{\hat{\rho}}=\mathrm{diag}(\bsig_{\hat{\rho}})$ and $\bSigma_{\pi}=\mathrm{diag}(\bsig_{\pi})$. For the diagonal approximation the objective turns into a sum of separable functions. 

\begin{equation}\label{sum_original_eq_ap}
\begin{split}
C_{\beta}(X,Y;\hat{\rho},\pi) &= \sum_i \frac{h_{i}}{2} \sigma_{i\hat{\rho}}+\sum_i\frac{\beta}{2\lambda}\frac{\sigma_{i\hat{\rho}}}{\sigma_{i\pi}}-\sum_i\frac{\beta}{2}+\sum_i\frac{\beta(\mu_{i{\hat{\rho}}}-\mu_{i\pi})^2}{2\lambda}\frac{1}{\sigma_{i\pi}}\\
&+\frac{\beta}{2}[\sum_i\ln(\lambda \sigma_{i\pi})-\sum_i\ln(\sigma_{i\hat{\rho}})]\\
&= \sum_i A_i \sigma_{i\hat{\rho}}+\sum_i B_i\frac{\sigma_{i\hat{\rho}}}{\sigma_{i\pi}}-\sum_i\frac{\beta}{2}+\sum_i C_i \frac{1}{\sigma_{i\pi}}+D_i[\sum_i\ln(\lambda \sigma_{i\pi})-\sum_i\ln(\sigma_{i\hat{\rho}})]\\
&= \sum_i[A_i \sigma_{i\hat{\rho}}+B_i\frac{\sigma_{i\hat{\rho}}}{\sigma_{i\pi}}-\frac{\beta}{2}+C_i \frac{1}{\sigma_{i\pi}}+D_i(\ln(\lambda \sigma_{i\pi})-\ln(\sigma_{i\hat{\rho}}))]\\
\end{split}
\end{equation}

where we have set $A_i=\frac{h_{i}}{2}$, $B_i=\frac{\beta}{2\lambda}$, $C_i=\frac{\beta(\mu_{i{\hat{\rho}}}-\mu_{i\pi})^2}{2\lambda}$, $D_i=\frac{\beta}{2}$. 

We take the derivatives of one of these functions with respect to $\sigma_{i\hat{\rho}},\sigma_{i\pi}$ and drop the indices $i$ for clarity

\begin{equation}
\frac{\partial C_{\beta}(X,Y;\hat{\rho},\pi)}{\partial\sigma_{\hat{\rho}}} = A+\frac{B}{\sigma_{\pi}}-\frac{D}{\sigma_{\hat{\rho}}}, \; \; \; \frac{\partial C_{\beta}(X,Y;\hat{\rho},\pi)}{\partial\sigma_{\pi}} = -\frac{B \sigma_{\hat{\rho}}}{\sigma_{\pi}^2}-\frac{C}{\sigma_{\pi}^2}+\frac{D}{\sigma_{\pi}}
\end{equation}
and
\begin{equation}
\frac{\partial C_{\beta}(X,Y;\hat{\rho},\pi)}{\partial^2\sigma_{\hat{\rho}}} = \frac{D}{\sigma_{\hat{\rho}}^2}, \; \; \; \frac{\partial C_{\beta}(X,Y;\hat{\rho},\pi)}{\partial^2\sigma_{\pi}} = 2(B \sigma_{\hat{\rho}}+C )\frac{1}{\sigma_{\pi}^3}-\frac{D}{\sigma_{\pi}^2}
\end{equation}
\begin{equation}
\frac{\partial C_{\beta}(X,Y;\hat{\rho},\pi)}{\partial\sigma_{\hat{\rho}}\partial\sigma_{\pi}} = -\frac{B}{\sigma_{\pi}^2}, \; \; \; \frac{\partial C_{\beta}(X,Y;\hat{\rho},\pi)}{\partial\sigma_{\pi}\partial\sigma_{\hat{\rho}}} = -\frac{B}{\sigma_{\pi}^2}
\end{equation}

We need to check whether the Hessian matrix is PSD so that the stationary point we found is a local minimum and the function is convex. We do that by calculating whether all principal minors of the Hessian are positive. 
\begin{equation}
\nabla^2C_{\beta}(\sigma_{\hat{\rho}},\sigma_{\pi}) = 
\begin{bmatrix}
    \frac{D}{\sigma_{\hat{\rho}}^2}       & -\frac{B}{\sigma_{\pi}^2}  \\
    -\frac{B}{\sigma_{\pi}^2}       & 2(B \sigma_{\hat{\rho}}+C )\frac{1}{\sigma_{\pi}^3}-\frac{D}{\sigma_{\pi}^2}  \\
\end{bmatrix}
\end{equation}
We see easily that $\text{det}(\frac{D}{\sigma_{\hat{\rho}}^2})>0$. While

\begin{equation}\label{det_first_dev_ap}
\begin{split}
\text{det}(\nabla^2C_{\beta}(\sigma_{\hat{\rho}},\sigma_{\pi})) &=  \frac{D}{\sigma_{\hat{\rho}}^2}\left(2(B \sigma_{\hat{\rho}}+C )\frac{1}{\sigma_{\pi}^3}-\frac{D}{\sigma_{\pi}^2}\right)-\frac{B^2}{\sigma_{\pi}^4}\\
&=\frac{1}{\sigma_{\hat{\rho}}^2 \sigma_{\pi}^4}\left(2CD\sigma_{\pi}-(D\sigma_{\pi}-B\sigma_{\hat{\rho}})^2 \right)\\
&=\left(\frac{1}{\sigma_{\hat{\rho}}^2 \sigma_{\pi}^4}\frac{\beta^2}{2}\right) \left(\frac{(\mu_{{\hat{\rho}}}-\mu_{\pi})^2}{\lambda}\sigma_{\pi}-\frac{1}{2}(\sigma_{\pi}-\frac{\sigma_{\hat{\rho}}}{\lambda})^2 \right)\\
\end{split}
\end{equation}

The determinant is not always positive and the function is not convex. We now check whether the sole stationary point is always a local minimum. We start by substituting $\sigma_{\hat{\rho}}^{\star}=\beta(h+\frac{\beta}{\lambda}\frac{1}{\sigma_{\pi}})^{-1}$ in the multiplicand of \eqref{det_first_dev_ap} as the multiplier is positive by definition

\begin{equation}\label{det_second_dev_ap}
\begin{split}
&\text{det}(\nabla^2C_{\beta}(\sigma_{\hat{\rho}}^{\star},\sigma_{\pi})) =\frac{1}{{\sigma_{\hat{\rho}}^{\star}}^2 \sigma_{\pi}^4}\frac{\beta^2}{2}\left(\frac{(\mu_{{\hat{\rho}}}-\mu_{\pi})^2}{\lambda}\sigma_{\pi}-\frac{1}{2}(\sigma_{\pi}-\frac{\beta}{\lambda}(h+\frac{\beta}{\lambda}\frac{1}{\sigma_{\pi}})^{-1})^2 \right)\\
& =\frac{1}{{\sigma_{\hat{\rho}}^{\star}}^2 \sigma_{\pi}^4}\frac{\beta^2}{2}\left(\frac{(\mu_{{\hat{\rho}}}-\mu_{\pi})^2}{\lambda}\sigma_{\pi}-\frac{1}{2}(\sigma_{\pi}-\frac{\beta}{\lambda}(\frac{\sigma_{\pi}\lambda}{h\lambda\sigma_{\pi}+\beta}))^2 \right)\\
& =\frac{1}{{\sigma_{\hat{\rho}}^{\star}}^2 \sigma_{\pi}^4}\frac{\beta^2}{2}\left(\frac{(\mu_{{\hat{\rho}}}-\mu_{\pi})^2}{\lambda}\sigma_{\pi}-\frac{\sigma_{\pi}^2}{2}(1-(\frac{\beta}{h\lambda\sigma_{\pi}+\beta}))^2 \right)\\
& =\frac{1}{{\sigma_{\hat{\rho}}^{\star}}^2 \sigma_{\pi}^3}\frac{\beta^2}{2}\left(\frac{(\mu_{{\hat{\rho}}}-\mu_{\pi})^2}{\lambda}-\frac{\sigma_{\pi}}{2}(\frac{h\lambda\sigma_{\pi}}{h\lambda\sigma_{\pi}+\beta})^2 \right)\\
& =\frac{1}{{\sigma_{\hat{\rho}}^{\star}}^2 \sigma_{\pi}^3}\frac{\beta^2}{2}\left(\frac{(\mu_{{\hat{\rho}}}-\mu_{\pi})^2}{\lambda}-\frac{\lambda^2 h^2\sigma_{\pi}^3}{2({h\lambda\sigma_{\pi}+\beta})^2} \right)\\
& = \frac{1}{{\sigma_{\hat{\rho}}^{\star}}^2\sigma_{\pi}^3 2 \lambda(h\lambda\sigma_{\pi}+\beta)^2}(2(\mu_{{\hat{\rho}}}-\mu_{\pi})^2(h\lambda\sigma_{\pi}+\beta)^2-\lambda^3h^2\sigma_{\pi}^3)\\
& = \frac{1}{{\sigma_{\hat{\rho}}^{\star}}^2 2 \lambda(h\lambda \Lambda_{{\pi}}^{-1}+\beta)^2}(2\Lambda_{{\pi}}(\mu_{{\hat{\rho}}}-\mu_{\pi})^2(h\lambda+\Lambda_{{\pi}}\beta)^2-\lambda^3h^2)\\
\end{split}
\end{equation}

Where we substituted $\sigma_{\pi} = \Lambda_{{\pi}}^{-1}$ as this will make the calculations easier. We now show a useful identity for $\Lambda_{{\pi}}^{\star}=\frac{\lambda}{2\beta}[\sqrt{h^2+\frac{4\beta h}{(\mu_{{\hat{\rho}}}-\mu_{\pi})^2}}-h]$

\begin{equation}\label{useful_identity_ap}
\begin{split}
{(\Lambda_{{\pi}}^{\star})}^2 &= \frac{\lambda^2}{4\beta^2}\left(h^2 +\frac{4\beta h}{(\mu_{{\hat{\rho}}}-\mu_{\pi})^2}-2h\sqrt{h^2 +\frac{4\beta h}{(\mu_{{\hat{\rho}}}-\mu_{\pi})^2}} + h^2 \right)\\ 
&=  \frac{\lambda^2}{4\beta^2}\left(2h\left(h-\sqrt{h^2 +\frac{4\beta h}{(\mu_{{\hat{\rho}}}-\mu_{\pi})^2}} \right) + \frac{4\beta h}{(\mu_{{\hat{\rho}}}-\mu_{\pi})^2}  \right)\\
&=  \frac{h\lambda}{\beta}\frac{\lambda}{2\beta}\left(\left(h-\sqrt{h^2 +\frac{4\beta h}{(\mu_{{\hat{\rho}}}-\mu_{\pi})^2}} \right) + \frac{2\beta}{(\mu_{{\hat{\rho}}}-\mu_{\pi})^2}  \right)\\
&=\frac{h\lambda}{\beta}\left( \frac{\lambda}{(\mu_{{\hat{\rho}}}-\mu_{\pi})^2} - \Lambda_{{\pi}}^{\star} \right)
\end{split}
\end{equation}

We substitute $\Lambda_{{\pi}} = \Lambda_{{\pi}}^{\star}$ in \eqref{det_second_dev_ap} and again develop only the multiplicand

\begin{equation}\label{det_third_dev_ap}
\begin{split}
&\text{det}(\nabla^2C_{\beta}(\sigma_{\hat{\rho}}^{\star},\sigma_{\pi}^{\star})) = \frac{1}{{\sigma_{\hat{\rho}}^{\star}}^2 2 \lambda(h\lambda {\Lambda_{{\pi}}^{\star}}^{-1}+\beta)^2}(2{\Lambda_{{\pi}}^{\star}}(\mu_{{\hat{\rho}}}-\mu_{\pi})^2(h\lambda+{\Lambda_{{\pi}}^{\star}}\beta)^2-\lambda^3h^2)\\
&= A(2{\Lambda_{{\pi}}^{\star}}(\mu_{{\hat{\rho}}}-\mu_{\pi})^2(h\lambda+{\Lambda_{{\pi}}^{\star}}\beta)^2-\lambda^3h^2)\\
&= A(2{\Lambda_{{\pi}}^{\star}}(\mu_{{\hat{\rho}}}-\mu_{\pi})^2(h^2\lambda^2+2h\lambda{\Lambda_{{\pi}}^{\star}}\beta+ {(\Lambda_{{\pi}}^{\star})}^2\beta^2)-\lambda^3h^2)\\
&= A(2{\Lambda_{{\pi}}^{\star}}(\mu_{{\hat{\rho}}}-\mu_{\pi})^2(h^2\lambda^2+2h\lambda{\Lambda_{{\pi}}^{\star}}\beta+ \frac{h\lambda}{\beta}\left( \frac{\lambda}{(\mu_{{\hat{\rho}}}-\mu_{\pi})^2} - \Lambda_{{\pi}}^{\star} \right)\beta^2)-\lambda^3h^2)\\
&= A(2{\Lambda_{{\pi}}^{\star}}(\mu_{{\hat{\rho}}}-\mu_{\pi})^2(h^2\lambda^2+h\lambda{\Lambda_{{\pi}}^{\star}}\beta+ \frac{ \beta \lambda^2 h } {(\mu_{{\hat{\rho}}}-\mu_{\pi})^2})-\lambda^3h^2)\\
&= A(2{\Lambda_{{\pi}}^{\star}}(\mu_{{\hat{\rho}}}-\mu_{\pi})^2(h^2\lambda^2+ \frac{ \beta \lambda^2 h } {(\mu_{{\hat{\rho}}}-\mu_{\pi})^2})+2{(\Lambda_{{\pi}}^{\star})^2}(\mu_{{\hat{\rho}}}-\mu_{\pi})^2h\lambda\beta-\lambda^3h^2)\\
&= A(2{\Lambda_{{\pi}}^{\star}}(\mu_{{\hat{\rho}}}-\mu_{\pi})^2(h^2\lambda^2+\frac{ \beta \lambda^2 h } {(\mu_{{\hat{\rho}}}-\mu_{\pi})^2})\\
&+2\frac{h\lambda}{\beta}\left( \frac{\lambda}{(\mu_{{\hat{\rho}}}-\mu_{\pi})^2} - \Lambda_{{\pi}}^{\star} \right)(\mu_{{\hat{\rho}}}-\mu_{\pi})^2h\lambda\beta-\lambda^3h^2)\\
&= A(2{\Lambda_{{\pi}}^{\star}}(\mu_{{\hat{\rho}}}-\mu_{\pi})^2(h^2\lambda^2+\frac{ \beta \lambda^2 h } {(\mu_{{\hat{\rho}}}-\mu_{\pi})^2})+2\lambda^3h^2-2h^2\lambda^2(\mu_{{\hat{\rho}}}-\mu_{\pi})^2{\Lambda_{{\pi}}^{\star}} -\lambda^3h^2)\\
&= A(2{\Lambda_{{\pi}}^{\star}}(\mu_{{\hat{\rho}}}-\mu_{\pi})^2(h^2\lambda^2+\frac{ \beta \lambda^2 h } {(\mu_{{\hat{\rho}}}-\mu_{\pi})^2})+\lambda^3h^2-2h^2\lambda^2(\mu_{{\hat{\rho}}}-\mu_{\pi})^2{\Lambda_{{\pi}}^{\star}})\\
&= A(2{\Lambda_{{\pi}}^{\star}}\beta \lambda^2 h+\lambda^3h^2)\\
&>0\\
\end{split}
\end{equation}

where we have set $A = \frac{1}{{\sigma_{\hat{\rho}}^{\star}}^2 2 \lambda(h\lambda {(\Lambda_{{\pi}}^{\star})}^{-1}+\beta)^2}>0$. We have used \eqref{useful_identity_ap} in lines 4 and 7.

Indeed the stationary point is a local minimum. We now show that there are no other local minima at the boundaries of the domain. From \eqref{sum_original_eq_ap} we see that we only need to evaluate expressions of the form $f(\sigma_{\hat{\rho}}) = \sigma_{\hat{\rho}}-\ln(\sigma_{\hat{\rho}})$ and $g(\sigma_{\pi}) = \frac{1}{\sigma_{\hat{\rho}}}+\ln(\sigma_{\hat{\rho}})$. By application of L'H\^opital's rule it's easy to show that 
\begin{equation}
\begin{split}
\lim_{\substack{\sigma_{\hat{\rho}}\to0 \\ \sigma_{\pi}=\text{ct}}}C_{\beta}(\sigma_{\hat{\rho}},\sigma_{\pi})&=\lim_{\substack{\sigma_{\hat{\rho}}\to+\infty \\ \sigma_{\pi}=\text{ct}}}C_{\beta}(\sigma_{\hat{\rho}},\sigma_{\pi})\\
&=\lim_{\substack{\sigma_{\hat{\rho}}=\text{ct} \\ \sigma_{\pi}\to0}}C_{\beta}(\sigma_{\hat{\rho}},\sigma_{\pi})
 =\lim_{\substack{\sigma_{\hat{\rho}}=\text{ct} \\ \sigma_{\pi}\to+\infty}}C_{\beta}(\sigma_{\hat{\rho}},\sigma_{\pi})=+\infty\\
\end{split}
\end{equation}

\end{proof}

\section*{D. Proof of Lemma 5.1}
\textbf{Preliminaries} We remind that a neural network transforms it's inputs $\ba_0=\bx$ to an output $f_{\btheta}(\bx)=\ba_l$ through a series of $l$ layers, each of which consists of a bank of units/neurons. The computation performed by each layer $i\in\{1,...,l\}$ is given as 

\begin{equation*}
\begin{split}
&\bs_i=\bW_i\ba_{i-1},\\
&\ba_i=\phi_i(\bs_{i}).\\
\end{split}
\end{equation*}

We also denote the vectorization of the weights as $\btheta=[\mathrm{vec}(\bW_0^{0,:})\mathrm{vec}(\bW_0^{1,:})\cdots\mathrm{vec}(\bW_0^{r,:})]$, where $\mathrm{vec}(\bW_i^{j,:})$ are the weights corresponding to layer $i$ and neuron $j$. We assume trained vectorized weights $\bmu_{\hat{\rho}i}$ and trained weights in matrix form $\bW_{\hat{\rho}i}$ for layer $i$. We will be adding bounded perturbations to the weights of each layer $i$ so that $||\bW_{i}-\bW_{\hat{\rho}i}||_F\leq C$. We will want to quantify the effect of these perturbations on the latent representations of the network. 

We then define $\bA_i=[\ba_i^0,\cdots,\ba_i^n]$, where $\ba_i^j$ is the unperturbed latent representation of sample $j$ at layer $i$, where $\bA_i$ is produced by the operation $\bA_i = \mathrm{rect}(\bW_{\hat{\rho}i}\bA_{i-1})$. We perturb only layer $i$ and define $\hat{\bA}_i$, as the representations resulting from the new perturbed matrix $\bW_i$, $\hat{\bA}_i = \mathrm{rect}(\bW_i\bA_{i-1})$. We then define $\tilde{\bA}_i$ as the representations at layer $i$ with accumulated error from layers $\leq i$. Similarly we can define the same quantities for the pre-activations $\bs_i^j$, we denote the corresponding matrices as $\hat{\bS}_i$ and $\tilde{\bS}_i$. 

We can then define the layerwise mean square error from perturbing only layer $i$
\begin{equation*}
\begin{split}
\hat{e}_i^2 &= (1/n)||\bA_i-\hat{\bA}_i||_F^2,\\
\hat{E}_i^2 &= (1/n)||\bS_i-\hat{\bS}_i||_F^2,\\
\end{split}
\end{equation*}
 as well as the accumulated mean square error
\begin{equation*}
\begin{split}
\tilde{e}_i^2 &= (1/n)||\bA_i-\tilde{\bA}_i||_F^2,\\
\tilde{E}_i^2 &= (1/n)||\bS_i-\tilde{\bS}_i||_F^2,\\
\end{split}
\end{equation*}
where the true representations are considered as constants. We make a simplifying assumption, assuming that the mean square error of our \emph{trained} classifier is 0. In this case we can set $\hat{\mathcal{L}}^{\ell_{\text{mse}}}_{X,Y}(f_{\btheta})\equiv\tilde{e}_l^2=(1/n)||\bA_l-\tilde{\bA}_l||_F^2$, as $\bA_l$ now correspond to the ground truth vectors. We can easily extend to the non-zero error case using the triangle inequality. 

These errors are difficult to analyze theoretically. As such we will make the useful assumption that they are well approximated by a \emph{quadratic}, which will make the analysis tractable. This assumption is quite strong and we do not claim that the approximation is tight. Furthermore Figure 3 of the main text does not directly apply in this setting; we will be dealing with the mean-square error instead of the categorical cross-entropy and we will be analyzing layerwise errors instead of the error at the output. At the same time our aim is only to derive a useful \emph{surrogate} objective. The empirical results in Section 5 provide evidence that the surrogate we propose is indeed useful in providing tighter bounds.

\textbf{Useful Lemmata} We prove the following Lemma which will be useful later. We first show that the mean square error at the output of a deep neural network can be decomposed as a sum of mean square errors for intermediate representations.

\begin{customlemma}{0.2}
Assuming layerwise perturbations that are bounded by a constant $||\bW_{i}-\bW_{\hat{\rho}i}||_F\leq C$, the accumulated mean square error $\tilde{e}_l^2$ at layer $l$ can be bounded as 
\begin{equation}
(1/n)||\bA_l-\tilde{\bA}_l||_F^2\leq\sum_{i=0}^lc_i(1/n)||\bA_i-\hat{\bA}_i||_F^2+\mathcal{O}(c^l)
\end{equation}
where $\forall i<l,\; c_i=\prod_{k=i+1}^l||\bW_k||_F^2$, $c_l=1$ and $c$ is some constant.
\end{customlemma}
\begin{proof}
We denote $\hat{a}_{i+1}$ a single element of $\hat{\ba}_{i+1}$ and $\bw_i^{\mathrm{T}}$ the corresponding row of $\bW_i$ where we drop the indices for individual samples and neurons for clarity. One can easily see through the properties of the rectifier function that
\begin{equation}
\begin{split}
\hat{a}_{i+1} &= \mathrm{rect}(\bw_{i+1}^{\mathrm{T}}\tilde{\ba}_{i}+\bw_{i+1}^{\mathrm{T}}(\ba_{i}-\tilde{\ba}_{i}))\\
&\leq \tilde{a}_{i+1}+\mathrm{rect}(\bw_{i+1}^{\mathrm{T}}(\ba_{i}-\tilde{\ba}_{i}))\\
&\leq \tilde{a}_{i+1}+|\bw_{i+1}^{\mathrm{T}}(\ba_{i}-\tilde{\ba}_{i})|\\
\end{split}
\end{equation}
Similarly we can obtain $\tilde{a}_{i+1}\leq \hat{a}_{i+1} +|\bw_{i+1}^{\mathrm{T}}(\ba_{i}-\tilde{\ba}_{i})|$ and therefore we can write
\begin{equation*}
|\tilde{a}_{i+1}-\hat{a}_{i+1}| \leq |\bw_{i+1}^{\mathrm{T}}(\ba_{i}-\tilde{\ba}_{i})|.
\end{equation*}
In matrix notation this becomes
\begin{equation*}
||\tilde{\bA}_{i+1}-\hat{\bA}_{i+1}||_F \leq ||\bW_{i+1}(\bA_{i}-\tilde{\bA}_{i})||_F\leq||\bW_{i+1}||_F||\tilde{\bA}_{i}-\bA_{i}||_F
\end{equation*}
By the triangle inequality we can then write
\begin{equation}
\begin{split}
\tilde{e}_{i+1}=(1/\sqrt{n})||\tilde{\bA}_{i+1}-\bA_{i+1}||_F&\leq(1/\sqrt{n})||\tilde{\bA}_{i+1}-\hat{\bA}_{i+1}||_F+(1/\sqrt{n})||\hat{\bA}_{i+1}-\bA_{i+1}||_F\\
&\leq(1/\sqrt{n})||\bW_{i+1}||_F||\tilde{\bA}_{i}-\bA_{i}||_F+(1/\sqrt{n})||\hat{\bA}_{i+1}-\bA_{i+1}||_F\\
&\leq \sum_{t=0}^i(\prod_{k=t+1}^{i+1}||\bW_{k}||_F||\hat{\bA}_{t}-\bA_{t}||_F)+(1/\sqrt{n})||\hat{\bA}_{i+1}-\bA_{i+1}||_F\\ 
&= \sum_{t=0}^i(\prod_{k=t+1}^{i+1}||\bW_{k}||_F\hat{e}_t)+\hat{e}_{i+1}\\
\end{split}
\end{equation}
If $||\bW_{i}-\bW_{\hat{\rho}i}||_F\leq C$, then the errors $\hat{e}_t=||\bA_t-\hat{\bA}_t||_F$ and also all terms $\prod_{k=t+1}^{i+1}||\bW_k||_F\hat{e}_t$ are bounded. We raise both sides to the power of 2. We get the desired terms as well as terms of the form $\prod_{k=a+1}^{i+1}||\bW_{k}||_F\prod_{k=b+1}^{i+1}||\bW_{k}||_F\hat{e}_a\hat{e}_b$ assuming that $||\bW_{i}||_F\leq\sqrt{c}$ we see that these are of the order $\mathcal{O}((\sqrt{c})^{2l})=\mathcal{O}(c^{l})$ and we get the desired result.
\end{proof}

In the following it will be useful to deal with the preactivations $\bs_i^j$ instead of the representations $\ba_i^j$ so as to avoid taking derivatives of the rectifier non-linearity. We will then find useful the following simple Lemma.
\begin{customlemma}{0.3}
Given the true preactivations $\bS_i$ and representations $\bA_i$, as well as the perturbed $\hat{\bS}_i$ and $\hat{\bA}_i$ for layer $i$ the following holds
\begin{equation}
(1/n)||\bA_i-\hat{\bA}_i||_F^2\leq(1/n)||\bS_i-\hat{\bS}_i||_F^2.
\end{equation}
\end{customlemma}
\begin{proof}
We assume $(\mathrm{rect}(x)-\mathrm{rect}(y))^2\leq(x-y)^2$ and check that it holds for different signs of $x,y$.
\end{proof}
We will now approximate the precativation error for each layer using a second order Taylor expansion. We prove the following.

\begin{customlemma}{0.4}
We apply a Taylor expansion of the layerwise preactivation error $\hat{E}_i^2(\btheta)$ of layer $i$, around a point $\bmu$. Given $j$ neurons and $n$ training samples, $\hat{E}_i^2(\btheta)$ can be approximated as 
\begin{equation}\label{layerwise_taylor}
\hat{E}_i^2(\btheta)=(1/n)||\bS_i-\hat{\bS}_i||_F^2 = \sum_j(\btheta_{ij}-\bmu_{ij})^T\bH_{i}(\btheta_{ij}-\bmu_{ij}) +\mathcal{O}(||\btheta_i-\bmu_{i}||^3).
\end{equation}
where $\bH_{i}=(1/n)\sum_{k=0}^n \ba_{i-1}^k{\ba_{i-1}^k}^T$.
\end{customlemma}
\begin{proof}
It will be easier to work with the vectorized weights per neuron $\btheta_{ij}$ directly. We note that the unperturbed representations $\bS_i$ are considered as constants, and get
\begin{equation}
\begin{split}
\frac{\partial\hat{E}_i^2}{\partial \btheta_{ij} } &=\frac{\partial}{\partial \btheta_{ij} }(1/n)||\hat{\bS}_i-\bS_i||_F^2\\
&=\frac{\partial}{\partial \btheta_{ij} }(1/n)||\bW_i\bA_{i-1}-\bS_i||_F^2\\
&=\frac{\partial}{\partial \btheta_{ij} }(1/n)\sum_{k=0}^n||\bW_i\ba_{i-1}^k-\bs_i^k||_2^2\\
&=\frac{\partial}{\partial \btheta_{ij} }(1/n)\sum_{k=0}^n\sum_{t=0}^r||\btheta_{it}^T\ba^k_{i-1}-s^k_{it}||^2_2\\
&=\frac{1}{n}\sum_{k=0}^n\sum_{t=0}^r\frac{\partial}{\partial \btheta_{ij} }||\btheta_{it}^T\ba^k_{i-1}-s^k_{it}||^2_2=\frac{2}{n}\sum_{k=0}^n(\btheta_{ij}^T\ba^k_{i-1}-s^k_{ij}){\ba^k_{i-1}}^T\\
\end{split}
\end{equation}
where in the third line we expand with respect to the samples and in the fourth line we expand with respect to each neuron. Then we can calculate the second order derivatives.
\begin{equation}
\frac{\partial^2 \hat{E}_i^2}{\partial^2 \btheta_{ij} } = \frac{\partial}{\partial \btheta_{ij}}\frac{2}{n}\sum_{k=0}^n(\btheta_{ij}^T\ba^k_{i-1}-s^k_{ij}){\ba^k_{i-1}}^T= \frac{2}{n}\sum_{k=0}^n \ba^k_{i-1}{\ba^k_{i-1}}^T.
\end{equation}
From the above, it is clear that the Hessian is block diagonal, with identical blocks for each neuron $j$. We can the approximate the layerwise error $\hat{e}_i^2$ using a second order Taylor expansion around a point $\bmu$ as
\begin{equation}
\begin{split}
\hat{E}_i^2 &= \frac{\partial \hat{E}_i^2}{\partial \btheta_{i} } (\btheta_{i}-\bmu_i)^T+\frac{1}{2}(\btheta_{i}-\bmu_i)^T \frac{\partial^2 \hat{E}_i^2}{\partial^2 \btheta_{i} } (\btheta_{i}-\bmu_i) +\mathcal{O}(||\btheta_i-\bmu_i||^3)\\
&= \sum_j [(\btheta_{ij}-\bmu_{ij})^T\sum_{k=0}^n\frac{1}{n}\ba^k_{i-1}{\ba^k_{i-1}}^T(\btheta_{ij}-\bmu_{ij})]+\mathcal{O}(||\btheta_i-\bmu_i||^3)\\
\end{split}
\end{equation}
where we assume that the derivatives with respect to the layer weights of order other than two are negligible. This is a strong but useful assumption to make, and one that will make the analysis tractable.
\end{proof}

We are now ready to prove our main lemma. 
\begin{customlemma}{5.1}
The differentiable surrogate objective
\begin{equation}\label{surrogate_mse_appendix}
\bE_{\btheta\sim\hat{\rho}(\btheta)}\hat{\mathcal{L}}^{\ell_{\text{mse}}}_{X,Y}(f_{\btheta})+\frac{1}{\beta n}(\mathrm{KL}(\hat{\rho}(\btheta)||\mathcal{N}(\bmu_{\pi},\lambda\boldsymbol{\mathrm{I}}))+\ln{\frac{1}{\delta}})
\end{equation}
, assuming that the layerwise derivatives of order other than 2 are negligible, has the following upper bound
\begin{equation}\label{surrogate_mse_upper_appendix}
\begin{split}
&\sum_{i,j} [\bE_{\bet_{ij} \sim \hat{\rho}_{ij}'(\btheta)}[\frac{1}{2}\bet_{ij}^T \bH_{i} \bet_{ij}] +\frac{1}{\beta n}\mathrm{KL}(\hat{\rho}_{ij}(\btheta)||\pi_{ij}(\btheta)]\\
&+\mathcal{O}(c^l)\\
\end{split}
\end{equation}
where $\hat{\rho}_{ij}(\btheta)=\mathcal{N}(\bmu_{\hat{\rho}ij},\bSigma_{\hat{\rho}ij})$, $\pi_{ij}(\btheta)=\mathcal{N}(\bmu_{\pi ij},\lambda\boldsymbol{\mathrm{I}})$, $\bH_{i}=(2/n)\sum_{k=0}^n \ba_{i-1}^k{\ba_{i-1}^k}^T$, are neuronwise posteriors, priors and Hessians.
\end{customlemma}
\begin{proof}
We assume that the prior $\pi(\btheta)$ and posterior $\hat{\rho}(\btheta)$ are block diagonal, with blocks corresponding to weights in each neuron.
\begin{equation}
\begin{split}
\bE_{\btheta\sim\hat{\rho}(\btheta)}[\hat{\mathcal{L}}^{\ell_{\text{mse}}}_{X,Y}(f_{\btheta})] &\leq \bE_{\btheta\sim\hat{\rho}(\btheta)}[\sum_{i=0}^lc_i(1/n)||\bA_i-\hat{\bA}_i||_F^2+\mathcal{O}(c^l)]\\
&\leq \bE_{\btheta\sim\hat{\rho}(\btheta)}[\sum_{i=0}^lc_i(1/n)||\bS_i-\hat{\bS}_i||_F^2+\mathcal{O}(c^l)]\\
&= \sum_{i=0}^l \bE_{\btheta\sim\hat{\rho}(\btheta)}[c_i] \bE_{\btheta\sim\hat{\rho}(\btheta)}[(1/n)||\bS_i-\hat{\bS}_i||_F^2]+\mathcal{O}(c^l)]\\
&\leq \sum_{i=0}^l c^* \bE_{\btheta\sim\hat{\rho}(\btheta)}[(1/n)||\bS_i-\hat{\bS}_i||_F^2]+\mathcal{O}(c^l)\\
&= \sum_{i=0}^l c^*\bE_{\bet_{ij}\sim\hat{\rho}_{ij}'(\btheta)}[\sum_j\bet_{ij}^T\bH_{i}\bet_{ij}]+\mathcal{O}(c^l)\\
&= \sum_{i,j} c^* \bE_{\bet_{ij}\sim\hat{\rho}_{ij}'(\btheta)}[\bet_{ij}^T\bH_{i}\bet_{ij}]+\mathcal{O}(c^l).\\
\end{split}
\end{equation}
In line 3 we used the fact that the constant $c_i$ for layer $i$ depends only on layers $k\geq i+1$, thus the two random variables are independent and the expectation operator is multiplicative. In line 4 we assume that the terms $c_i=\prod_{k=i+1}^l||\bW_k||_F^2$ are upper bounded by the constant $c^*$. This is reasonable as in practice we will be adding Gaussian noise with bounded variance to the layer weights. In line 5 we approximate the error $\hat{E}_i^2=(1/n)||\bS_i-\hat{\bS}_i||_F^2$ using \eqref{layerwise_taylor} at point $\mu_{\hat{\rho}}$ which is the mean of the posterior $\hat{\rho}(\btheta)$, then we use that $\hat{\rho}'(\btheta)$ is a centered version of $\hat{\rho}(\btheta)$. We finally assume that the term $\mathcal{O}(c^l)$ dominates the remainders from the Taylor expansion.

We then absorb the constant $c^*$ in the hyperparameter $\beta$. By noting that the KL divergence of block-diagonal Gaussians can be decomposed as $\mathrm{KL}(\mathcal{N}(\hat{\rho}(\btheta)||\pi(\btheta))=\sum_{ij}\mathrm{KL}(\mathcal{N}(\hat{\rho}_{ij}(\btheta)||\pi_{ij}(\btheta))$  we get the desired result. 
\end{proof}

\textbf{Importantly} we don't require that the deep neural network was trained using the mean square error. Rather we can optimize \eqref{surrogate_mse_upper_appendix} for any network and assume that it's representations remain close based on the mean square error. Our experiments however show that optimizing \eqref{surrogate_mse_upper_appendix} is also a good surrogate for keeping the 01-error small.

\section*{E. Experimental Setup}
Experiments for Variational Inference were performed on NVIDIA Tesla K40c GPU.
All other experiments were performed on an NVIDIA GEFORCE GTX 1080 GPU. The libraries used were Tensoflow 1.15.0 \cite{tensorflow2015-whitepaper}, Keras 2.2.4 \cite{chollet2015keras} and Tensorflow-Probability 0.8.0 \cite{dillon2017tensorflow}.

When training the original deterministic classifiers, for the MNIST architectures we used the Keras implementation SGD with a learning rate of $0.01$, momentum value of $0.9$ and exponential decay with decay factor $0.001$. For CIFAR architectures we used the Keras implementation of Adam with a learning rate of $0.001$, $\beta_1=0.9$, $\beta_2=0.999$, decay value of $0.00005$ and the default value for the epsilon parameter. We used the softmax activation as well as the categorical cross-entropy in both cases. MNIST architectures were trained for 10 epochs while CIFAR architectures where trained for 200 epochs, which was sufficient for the training loss to stop decreasing.

When optimizing the posterior distributions centered at the deterministic classifier we used a grid search over $\beta$ and/or $\lambda$ where appropriate, with limits specified in the following tables. The computational time reported refers to the total time required to compute the plots in the main text for each setup, including computing the posterior and/or prior distributions as well as sampling $m$ number of samples for estimating the expected empirical risk of the stochastic classifier.

\textbf{MNIST.} We report the following values for the MNIST experiments.

\begin{tabularx}{0.8\textwidth} { 
  | >{\raggedright\arraybackslash}X 
  | >{\centering\arraybackslash}X 
  | >{\centering\arraybackslash}X 
  | >{\raggedleft\arraybackslash}X | }
 \hline
 Experiment & $\beta$ & $\lambda$ & Time \\
 \hline
 MNIST Is@0  & -  & [0.031,0.3]  & 14h \\
\hline
 MNIST Is@Init  & -  & [0.031,0.3]  & 14h \\
\hline
 MNIST VI  & [1,5]  & [0.03,0.1]  & 11h \\
\hline
 MNIST Post   & [0.001,0.07]  & [0.00005,0.01]  & 33h \\
\hline
 MNIST Post+Prior   & [0.000007,0.001]  & -  & 10h \\
\hline
 MNIST sK-FAC   & [0.001,0.02]  & [0.001,0.1]  & 33h \\
\hline
\end{tabularx}

The $\beta$ and $\lambda$ ranges are identical for MNIST10, MNIST5, MNIST2 while computation times are of the same order of magnitude.

\textbf{CIFAR.} We report the following values for the CIFAR experiments.

\begin{tabularx}{0.8\textwidth} { 
  | >{\raggedright\arraybackslash}X 
  | >{\centering\arraybackslash}X 
  | >{\centering\arraybackslash}X 
  | >{\raggedleft\arraybackslash}X | }
 \hline
 Experiment & $\beta$ & $\lambda$ & Time \\
 \hline
 CIFAR Is@0  & -  & [0.031,0.3]  & 15h \\
\hline
 CIFAR Is@Init  & -  & [0.031,0.3]  & 15h \\
\hline
 CIFAR VI  & [1,2]  & [0.1,0.3]  & 10h \\
\hline
 CIFAR Post   & [0.001,0.1]  & [0.001,0.1]  & 32h \\
\hline
 CIFAR Post+Prior   & [0.0001,0.001]  & -  & 11h \\
\hline
 CIFAR sK-FAC   & -  & -  & - \\
\hline
\end{tabularx}

The $\beta$ and $\lambda$ ranges are identical for CIFAR10, CIFAR5, CIFAR2 while computation times are of the same order of magnitude.

For the Variational Inference experiments we used the Adam \cite{kingma2014adam} optimizer with a learning rate of $1e-1$ for 5 epochs of training. For efficient inference we used the Tensorflow-Probability \cite{dillon2017tensorflow} implementation of the Flipout \cite{wen2018flipout} estimator.

\section*{F. Notes on PAC-Bayes}
We note here some important differences between the PAC-Bayesian setting and the standard Bayesian treatment of deep neural networks, as there are some important overlaps in the terms used. 

First, while PAC-Bayes refers to a ``posterior" $\hat{\rho}$ this distribution is not required to be a posterior in the Bayesian sense. On the contrary it can be chosen to be \emph{any} distribution. As such we are free to model $\hat{\rho}$ using different distributions centered on the deterministic neural networks, decoupled from how we trained the original deterministic network. In particular in Section 5 we can minimize the mean square error surrogate from Lemma 5.1. even though the deterministic networks are trained using the categorical cross-entropy loss. 

Second, as noted in the main text the prior $\pi$ in PAC-Bayes has to be independent of the training set but can depend on the data distribution.

\end{document}